\pdfoutput=1

\documentclass[11pt]{article}

\usepackage{EMNLP2022}

\usepackage{times}
\usepackage{latexsym}

\usepackage[T1]{fontenc}
\usepackage[utf8]{inputenc}
\usepackage{microtype}
\usepackage{inconsolata}
\usepackage{enumitem,kantlipsum}
\usepackage{graphicx}
\usepackage{caption}
\usepackage{amsmath}
\usepackage{amssymb}
\usepackage{pgfplots}
\pgfplotsset{compat=newest}
\usepackage{comment}
\usepackage{booktabs}
\usepackage{hyperref}
\usepackage[overload]{empheq}
\usepackage{makecell}
\usepackage{xcolor,colortbl}
\usepackage[ruled,vlined]{algorithm2e}
\usepackage{array,multirow}
\usepackage{comment}
\usepackage{pifont}
\usepackage{url}
\usepackage{siunitx}

\setlist[itemize]{leftmargin=*}

\newcommand{\STAB}[1]{\begin{tabular}{@{}c@{}}#1\end{tabular}}

\newcommand{\cmark}{\ding{51}}%
\newcommand{\xmark}{\ding{55}}%

\newcommand{\ella}[1]{{\textit{\color{blue}{#1}}}}

\usepackage[T1]{fontenc}

\usepackage[utf8]{inputenc}

\usepackage{microtype}

\title{Predicting Question-Answering Performance of Large Language Models through Semantic Consistency}

\author{
	Ella Rabinovich\hspace{3cm}
	Samuel Ackerman\hspace{3cm}
	Orna Raz
	\vspace{0.1cm} \\
	\textbf{Eitan Farchi}\hspace{4cm} 
	\textbf{Ateret Anaby-Tavor} 
	\vspace{0.15cm} \\
	IBM Research \\
	\texttt{\{ella.rabinovich1, samuel.ackerman\}@ibm.com} \\
	\texttt{\{ornar, farchi, atereta\}@il.ibm.com}
}

\begin{document}
\maketitle
\begin{abstract}
Semantic consistency of a language model is broadly defined as the model's ability to produce semantically-equivalent outputs, given semantically-equivalent inputs. We address the task of assessing question-answering (QA) semantic consistency of contemporary large language models (LLMs) by manually creating a benchmark dataset with high-quality paraphrases for factual questions, and release the dataset to the community.

We further combine the semantic consistency metric with additional measurements suggested in prior work as correlating with LLM QA accuracy, for building and evaluating a framework for factual QA reference-less performance prediction -- predicting the likelihood of a language model to accurately answer a question. Evaluating the framework on five contemporary LLMs, we demonstrate encouraging, significantly outperforming baselines, results.
\end{abstract}

\section{Introduction}
\label{sec:introduction}

Consistency of a model is broadly defined as the invariance of its behavior under meaning-preserving variations of its input \citep{elazar2021measuring, raj2022measuring}. Clearly, consistency is a highly desirable property of large language models, increasing their safety, robustness and trustworthiness. Here we address the question of \textit{factual} consistency of LLMs in the context of open-domain zero-shot factual question answering. As a concrete example, a consistent model will produce the same answer for the set of questions \{``What is Stevie Cameron's occupation?'', ``What job does Stevie Cameron do?'', ``What does Stevie Cameron earn a living as?''\}. A model's consistency metric is defined to be agnostic to answers' accuracy \citep{elazar2021measuring}, meaning that semantically-equivalent (possibly incorrect) outputs are qualified as consistent. As such, while the correct answer to the questions above is ``journalist'', three other identical answers (e.g., ``politician'') will score as perfectly consistent.

Semantic consistency of masked language models (MLMs) has been studied by \citet{elazar2021measuring}, who inspected masked tokens as predicted by encoder models, for alternations of word tuples, using a dataset of factual statements and their crowd-sourced paraphrases, specifically tailored for working with MLMs. \citet{raj2022measuring} evaluated semantic consistency of decoder models for the task of non-factual question answering, experimenting with a range of consistency metrics. The authors automatically generated paraphrases for questions in the TruthfulQA dataset \citep{lin2022truthfulqa}, and scored a model's consistency as its robustness to paraphrases. However, the sub-optimal quality of automatic paraphrases, along with open and often lengthy nature of answers to questions,\footnote{Example questions: "What happens if an earthworm is cut in half?", "Do you only own a gun for self defense?"} as well as multiple (occasionally semantically diverse) reference answers, challenge benchmarking of LLMs' QA consistency using TruthfulQA.

A benchmark dataset for measuring the robustness of LLMs to paraphrases in the context of factual QA should satisfy two  desirable properties: (1) strictly semantically-equivalent question paraphrases, and (2) questions that call for single short (possibly multi-word) answer, facilitating accurate evaluation. Using the recently introduced PopQA dataset with over 14K factual questions \citep{mallen2023not}, we create its carefully curated extended version---PopQA-TP (PopQA templated paraphrases)---where 3--10 manually-created alternations were appended for each original question. The final dataset comprises over 118K questions, while preserving metadata (e.g., reference answers) from the original PopQA. We further use this dataset for benchmarking factual semantic consistency of multiple encoder-decoder and decoder-only LLMs. The dataset is made available for the community at \url{https://huggingface.co/datasets/ibm/popqa-tp}.

We next demonstrate that robustness to question paraphrases correlates with a model's answer correctness for the given question. Practically, this finding means that semantic consistency score is predictive of a model accuracy. Combining this predictor with additional metrics suggested in prior work as correlating with LLM QA correctness, we perform a comprehensive regression analysis of the predictive power of various metrics on the model's accuracy, as well as interactions between those metrics. Moreover, we show that the developed framework can be used for predicting the likelihood of a language model to accurately answer a factual question. Collectively, these results pave the way for the extremely challenging, yet highly important, task of \textit{question-answering performance prediction}, a reference-less evaluation of QA performance, in the absence of ground-truth answers.

The contribution of this work is, therefore, twofold: First, we introduce and release a large extension of the PopQA dataset (PopQA-TP), with high-quality paraphrases, that can be used for benchmarking QA semantic consistency of LLMs. Second, we develop a prototype model for QA performance prediction, allowing for comparative analysis of various metrics, and demonstrating predictive power much higher than baselines.

\section{Dataset}
\label{sec:dataset}

Benchmarking semantic consistency of LLMs requires high quality question alternations, eliminating possible confounds that stem from issues in automatic paraphrase generation. Despite the immense advances in paraphrasing models during the past few years (e.g., \citealt{bandel2002quality, raj2022measuring, rahamim2023text}), automatic tools still occasionally produce paraphrases that are not meaning-preserving (e.g., ``Who is the vocalist of `Perfect'?'' for the original question ``Who is the composer of `Perfect'?''), incomplete (e.g., ``Who is the vocalist of `Perfect'? Shape of You''), or violate, albeit infrequently, grammatical rules (e.g., ``Tap water's safe drinking?" as a paraphrase of ``Is tap water safe to drink?"). Aiming at a high-quality benchmark dataset, we opted to manually construct paraphrase templates specific to each question category in PopQA, as detailed below.

\subsection{Paraphrase Templates Creation}

Each question $q\in\textrm{PopQA}$ is formed by substituting a single-entity subject into a question template that is fixed for each category.  For instance, the \textit{occupation} and \textit{religion} templates are ``What is <subject>'s occupation?" and ``What is the religion of <subject>?", respectively.  These fixed templates are sometimes grammatically awkward depending on the type of subject, for instance for the \textit{religion} category subject `Assumption of Mary'.

We create the paraphrase question dataset by manually creating multiple paraphrase templates specific to each category, and substituting the subject of each $q$ in PopQA into each template, yielding a set of paraphrases denoted by $P(q)$.  Thus, each question in a given category has the same number of paraphrases.  We name the resulting dataset PopQA-TP (PopQA templated paraphrases), which thus consists of $(P(q){+}\{q\}\colon q{\in}\textrm{PopQA})$, that is, the original questions and their paraphrases. 

Table~\ref{tbl:popqa-expanded-stats} shows summary statistics of the number of questions, by category and overall, for both the original PopQA and our PopQA-TP datasets. Examples of original questions and paraphrases in PopQA-TP 
are reported in Table~\ref{tbl:popqa-expanded-examples}.

\begin{table}[hbt]
\centering
\resizebox{\columnwidth}{!}{
\begin{tabular}{l|rrr}
category & \# Q & \# Q alternatives  &  total \# Q \\ \hline
author & 1514 & 6 & 9084 \\
capital & 645 & 6 & 4515 \\
capital of & 363 & 3 & 1452 \\
color & 34 & 5 & 204 \\
composer & 978 & 5 & 5868 \\
country & 838 & 9 & 8380 \\
director & 1999 & 10 & 21989 \\
father & 570 & 4 & 2850 \\
genre & 1619 & 6 & 11333 \\
mother & 187 & 5 & 1122\\
occupation & 532 & 5 & 3192\\
place of birth & 584 & 6 & 4088\\
producer & 1520 & 10 & 16720 \\
religion & 338 & 5 & 2028 \\
screenwriter & 1999 & 10 & 21989\\
sport & 547 & 6 & 3829\\
\hline
total & 14267 & & 118643\\
\end{tabular}
}
\caption{Dataset summary statistics, for each category label in PopQA.  Column `\#Q' shows the number of original questions, one per subject, in PopQA; column `\#Q alternatives' is the number of template paraphrase for each question in that category, in our PopQA-TP dataset; `total \# Q' is the resulting number of questions in PopQA-TP, which is $(1+ (\textrm{\#Q alternatives)}) \times (\textrm{\# Q})$.}
\label{tbl:popqa-expanded-stats}
\end{table}

\begin{table}[hbt]
\centering
\resizebox{\columnwidth}{!}{
\begin{tabular}{l}
question \\ \hline
What genre is Avatar: The Last Airbender? \\ \hline
What type of work is Avatar: The Last Airbender? \\
Fans of what genre would like Avatar: The Last Airbender? \\
What genre does Avatar: The Last Airbender belong to? \\
What genre is "Avatar: The Last Airbender"? \\
What genre is Avatar: The Last Airbender associated with? \\
Avatar: The Last Airbender is associated with what genre? \\ \hline \hline
What is Shozaburo Nakamura's occupation? \\ \hline
What is the occupation of Shozaburo Nakamura? \\
What kind of work does Shozaburo Nakamura do? \\
What does Shozaburo Nakamura earn a living as? \\
What job does Shozaburo Nakamura do? \\
What is Shozaburo Nakamura's job? \\
\end{tabular}
}
\caption{Example set of question paraphrases in PopQA-TP for the \textit{genre} and \textit{occupation} categories. The first question in each paraphrase grouping is the original question from PopQA.}
\label{tbl:popqa-expanded-examples}
\end{table}

Some PopQA question categories contain subjects of the same underlying type, while in others the type may vary.  For instance, subjects of \textit{occupation} questions are all persons, and in \textit{capital of} they are all states, provinces, or countries, etc.  In \textit{religion}, some are persons (e.g., Rumi or Paul, but also people like Bertrand Russell who were not religious leaders), ethnic or national groups (e.g., Swedes, Arabs), institutions (e.g., Boston College), or miscellaneous topics (e.g., saint, Bourbon Restoration, Assumption of Mary).  For some subjects, thus, it would be more grammatical to phrase the religion question as what religion the subject `follows', and in for others which religion the subject is `associated with'.  Note that this awkwardness is inherent in the original PopQA, and so our paraphrase templates are designed to span the possible meanings.  Nevertheless, we expect a good model to answer these questions intelligently and not be stumped by slight grammatical awkwardness.



Throughout the work, we obtain text vector embeddings using the SentenceTransformer (ST) encoder \citep{reimers2019sentence}.  The quality of paraphrases of $q$ can thus be assessed by the average cosine similarity between the embeddings of each paraphrase and $q$.  Calculating the average paraphrase quality for each question category, and averaging across categories, we obtain a high value of 0.914; this shows that the templated paraphrases are sufficiently similar to the original questions.

\section{Benchmarking Semantic Consistency}
\label{sec:benchmarking-experiments}

We next use PopQA-TP, our dataset of manually-constructed paraphrase templates for assessing the semantic consistency of multiple contemporary LLMs. We report both models' accuracy (the ratio of correct answers to questions), as well as their consistency (robustness to question alternations), and further develop hypothesis about the correlation of semantic consistency and correctness.

\subsection{Experimental Setup}
\label{ssec:models}
We experiment with several openly-available encoder-decoder and decoder-only contemporary LLMs, that have been proven effective in multiple generative tasks: Google Research's Flan-T5-XXL (11B; \citealp{chung2022scaling}) and Flan-UL2 (20B; \citealp{flan_ul2}), BigScience Workshop's MT0-XXL (13B; \citealp{bigscience_mt0_xxl}), EleutherAI's GPT-NeoX (20B; \citealp{eleutherai_gpt_neox}) and Mosaic ML, Inc.'s MPT-Instruct2 (7B; \citealp{mosaicml_mpt}).

Each question in PopQA-TP is queried to each model in greedy decoding mode, i.e., no sampling is allowed. Following previous studies \citep{raj2022measuring}, for the decoder-only models, the prompt is formatted using the input query template \texttt{Question:{<*>}\textbackslash n Answer:}, while for the encoder-decoder models, it is submitted as-is. The GPT-NeoX and MPT-Instruct2 models often generated multi-sentence answers; 
in these cases, only the first sentence was used for evaluation.

\subsection{Semantic Consistency -- Metrics}
\label{ssec:consistency-metrics}
Semantic consistency of a language model is broadly defined as the model's ability to produce semantically-equivalent outputs, given semantically-equivalent inputs \citep{elazar2021measuring, jang2021accurate, zhou2022prompt}. The precise approach to consistency assessment may, however, vary according to the characteristics of the generated text. Here we distinguish between free-form (possibly long) answers to open questions, and short, often single-word, factoid answers.

\paragraph{Semantic Consistency of Free-form Answers}
In the context of open-domain zero-shot QA, \citet{raj2022measuring} quantify the equivalence of a model's answers to semantically-equivalent paraphrases of the same question. The authors show, among others, that semantic equivalence of relatively long (sentence- or short paragraph-length) answers, is most reliably quantified by means of measuring \textit{lexical entailment} between pairs of answers. In particular, they demonstrate higher correlation of this metric to human judgements, than e.g., using pairwise cosine similarity between answers' dense representations. As a concrete example, consider two answers for rephrases of the question "What are the benefits of eating an apple a day?" (expanded TruthfulQA, \citealp{raj2022measuring}):

(1) \textit{Apples are a delicious and nutritious fruit that offer a range of health benefits when consumed regularly.} (2) \textit{Apples are a popular and healthy food that provide numerous benefits.}

While the second answer could be reasonably entailed from the first one (and vise versa), cosine similarity between the two embeddings might not be very indicative of their (rough) equivalence due to the relatively high lexical distinction.

\paragraph{Semantic Consistency of Factoid Answers}
Contrary to questions that call for a (possibly long) free-form answer, PopQA, and its paraphrase-extended version, require short, single- or a few-word answers, that constitute a less-natural fit for the task of lexical entailment. Alternatively, cosine similarity of answer embeddings provides a more reliable similarity score for very short utterances. 
As an example, semantic consistency rating of two answers to the question "What is {\textless person\textgreater}'s occupation?" (PopQA, \citealt{mallen2023not}), with a SOTA NLI model\footnote{\url{https://huggingface.co/microsoft/deberta-xlarge-mnli}} and cosine similarity is reported in Table~\ref{tbl:nli-cosine-scores}:

\begin{table}[hbt]
\centering
\resizebox{\columnwidth}{!}{
\begin{tabular}{l|l|c|c}
answer 1 & answer 2 & NLI & cosine \\ \hline
actress & actress & 0.927 & 1.00 \\
architect & architect & 0.876 & 1.00 \\
politician & german politician & 0.075 & 0.70 \\
german politician & politician & 0.588 & 0.70 \\
\end{tabular}
}
\caption{NLI and cosine similarity scores of two answers to rephrases of the same question. Note the NLI score distinction between the two ``politician'' examples due to the inherently asymmetric nature of lexical entailment, as well as differences for ``actress'' and ``architect''.}
\label{tbl:nli-cosine-scores}
\end{table}


\subsection{Experimental Results}
We next present the results of LLMs correctness and semantic consistency, using PopQA-TP.

\subsubsection{Correctness}
\begin{figure*}[h!]
\centering
\resizebox{1.0\textwidth}{!}{
\includegraphics{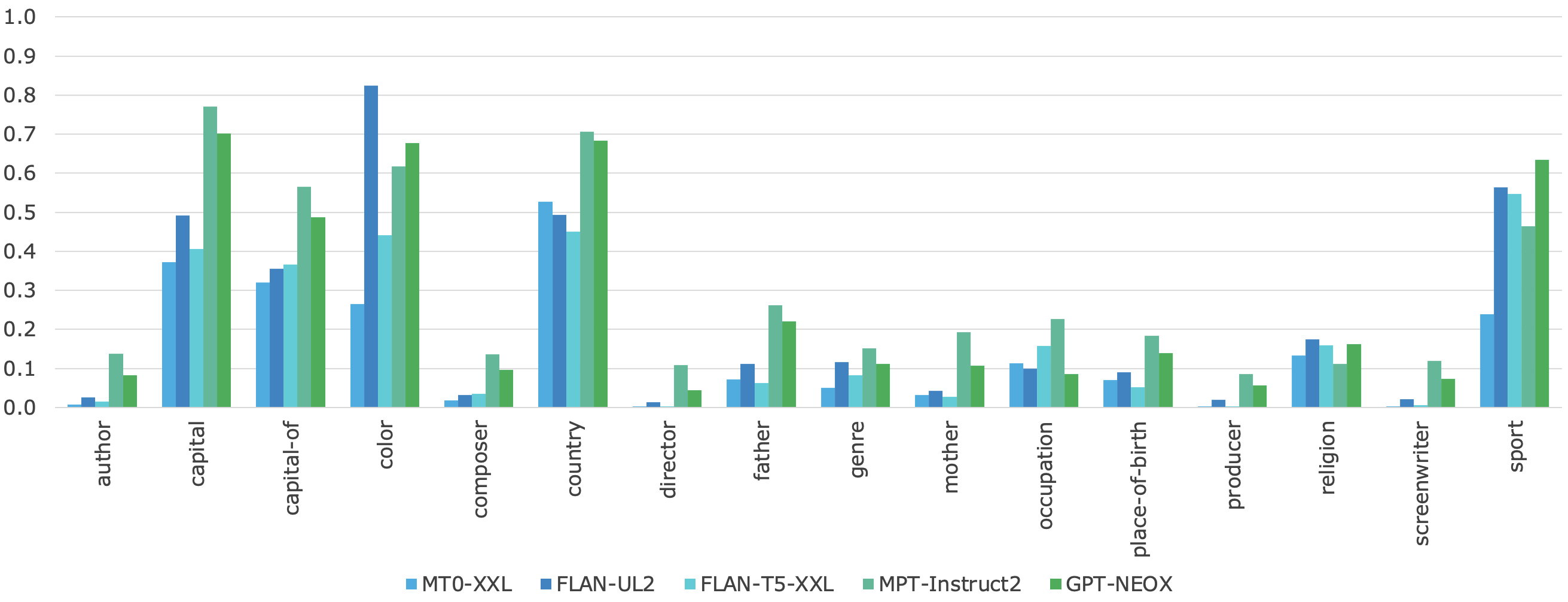}
}
\caption{Mean LLMs' correctness on questions in the PopQA dataset \cite{mallen2023not}, by category. Blue shades denote encoder-decoder models, green -- decoder-only.}
\label{fig:models-correctness}
\end{figure*}

Following \citet{mallen2023not}, we consider a question answered correctly if a substring of the generated text is an exact string match to one of the gold answers (e.g., a generated answer of "film director" matches "director"). Figure~\ref{fig:models-correctness} presents the mean correctness results for the five models, split by category. Evidently, some categories are systematically easier than others, e.g., \textit{color} and \textit{sport}, while others pose challenge across the board, e.g., \textit{author} and \textit{director}. This result can be partly attributed to the more restricted space of plausible answers to former categories (there is only a limited set of color names), compared to the infinitely large space of person names for the latter. Notably, the two decoder-only models---MPT-Instruct2 (accuracy of 0.224) and GPT-NeoX (accuracy of 0.184)---perform better than their encoder-decoder counterparts, on average, across categories.

\subsubsection{Semantic Consistency}
Internal semantic consistency of a set of (possibly non-unique) texts $\mathcal{T}{=}\{t_1,t_2,\dots\}$ can be calculated by the mean pairwise cosine similarity of their respective embedding vectors $\{e_1,e_2,\dots\}$, which ranges from 0 to 1. Formally:
\begin{equation}
\textrm{int\_sim}(\mathcal{T}){=}\frac{1}{{|\mathcal{T}| \choose 2}}\sum_{i=1}^{|\mathcal{T}|-1}\sum_{j=i+1}^{|\mathcal{T}|} \textrm{cosine}(e_i,e_j)
\end{equation}

Given $\mathcal{A}$, the set of generated answers to $q$ and paraphrases $P(q)$, we define the semantic consistency of $\mathcal{A}$ as $\texttt{SCons}(q){=}\textrm{int\_sim}(\mathcal{A}){\in}[0,1]$.  

\begin{figure*}[h!]
\centering
\resizebox{1.0\textwidth}{!}{
\includegraphics{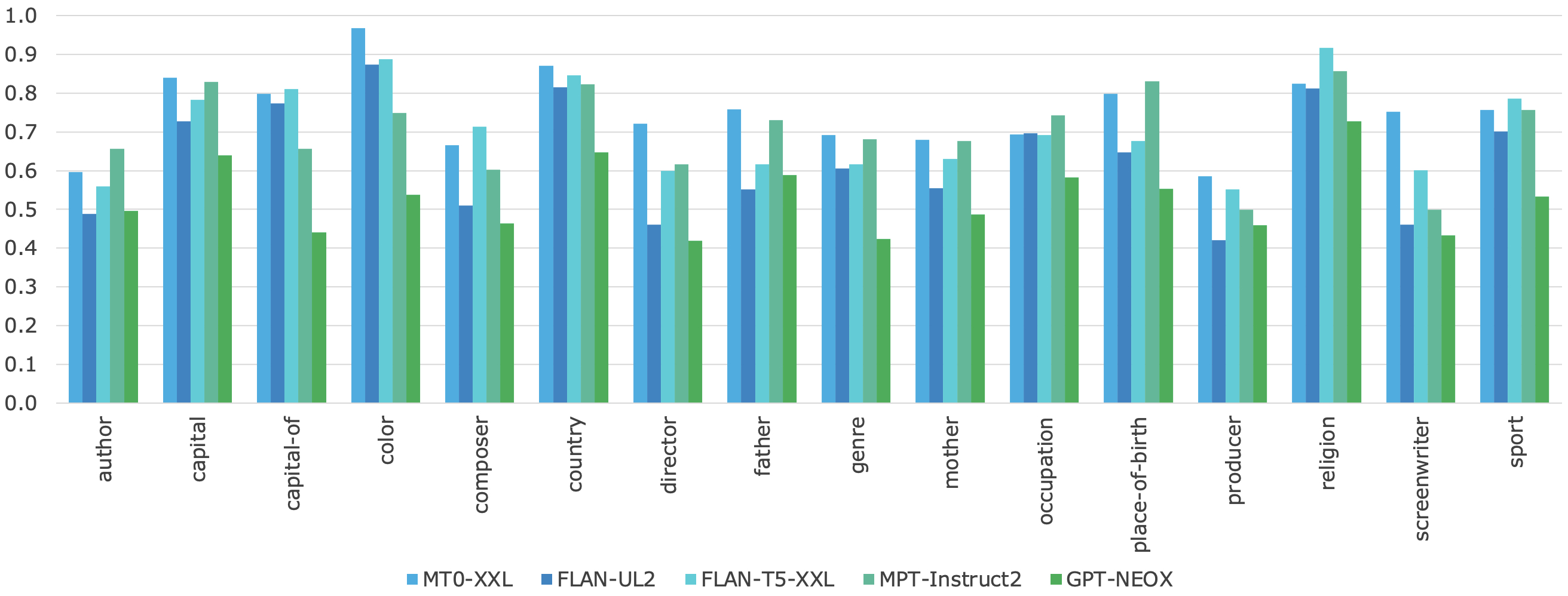}
}
\caption{Mean LLMs' consistency on questions in the PopQA dataset \cite{mallen2023not} and their paraphrases (PopQA-TP, this work), by category. Blue shades denote encoder-decoder models, green -- decoder-only.}
\label{fig:models-consistency}
\end{figure*}

\begin{figure}[h!]
\centering
\includegraphics[width=\columnwidth]{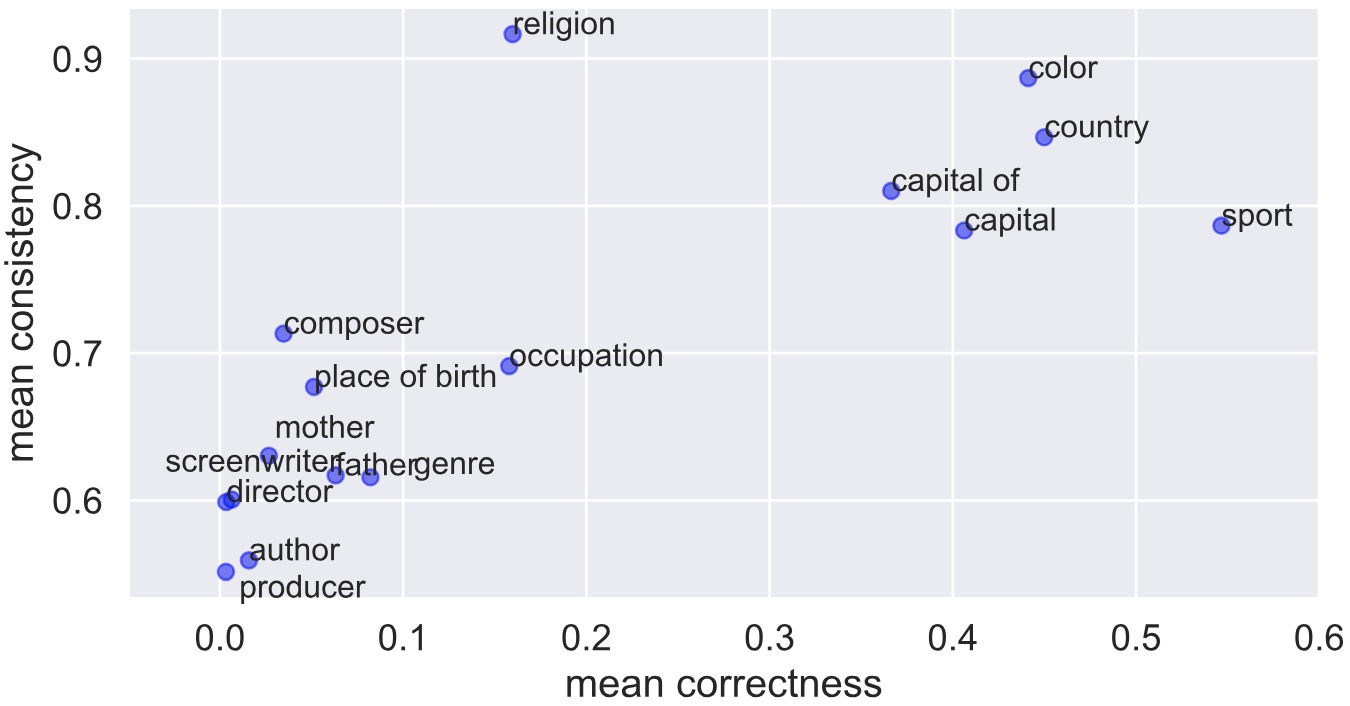}
\caption{Scatterplot of mean in-category answer correctness and consistency (as depicted in Figures~\ref{fig:models-correctness} and ~\ref{fig:models-consistency}) for the Flan-T5-XXL model. The evident positive correlation supports the intuition that semantic consistency has a predictive power on an LLM QA accuracy.
}
\label{fig:models-scatterplot}
\end{figure}

Figure~\ref{fig:models-consistency} presents results of mean answer semantic consistency computation, by question category. Consistency values vary in the [0.4, 0.9] range, with some (albeit lower) deviation across categories. Similarly to correctness, the relatively high consistency values in \textit{capital}, \textit{color}, \textit{country}, \textit{religion}, and \textit{sport} can be attributed to the more restricted space of plausible answers, compared to other categories.  Figure~\ref{fig:models-scatterplot} shows a scatterplot of the mean category correctness and consistency for the Flan-T5-XXL as a representative example of the models.  Across categories, answer correctness and consistency are positively correlated.  Across all models considered, the religion category is an outlier among the categories above with restricted answer space, in that these questions had relatively low correctness but high consistency.


Contrary to correctness results, here encoder-decoder LLMs (MT0-XXL, Flan-UL2 and Flan-T5-XXL) outperform decoder-only models.

\section{QA Performance Prediction}
\label{sec:performance-prediction}


We next define and address the task of factual question-answering performance prediction. 
Here we rely on some parallels to the task of \textit{query performance prediction} (QPP) in IR (search) systems -- an established research area \citep{zhou2007query, carmel2012query, raiber2014query, faggioli2023query}. QPP is defined as the assessment of the retrieval quality of a search system for a query, without relevance judgments. Core differences exist between IR and LLM-based systems used for the task of open-domain factual QA; yet, we address a conceptually similar task: assessment of a system's potential answer quality (that is manifested by its correctness) for a question, without relying on ground-truth answers.

Casting the task as a classification scenario, we train a logistic regression model, where several regressors---variables proven to correlate with LLMs correctness---carry over predictive power on the outcome variable: the model's likelihood to produce a correct answer for a given question.

\subsection{Predictor Variables}
\label{ssec:preictors}


\subsubsection{Question Subject Popularity (\texttt{SPop})
\label{ssec:subject_popularity}}
\citet{mallen2023not} hypothesize that factual knowledge that is less frequently discussed on the web may not be well memorized by LLMs. Given a question that can be modeled by the \{subject, relationship, object\} triple, e.g., ``What is the \textit{capital of} (R) \textit{Louisiana} (S)?'', the authors approximate its subject's popularity by the mean number of monthly views of the corresponding Wikipedia page. The answer---``Baton Rouge''---is scored by popularity in a similar way, but we refrain from using this score for our predictive analysis, since it is unknown in a realistic QA setup.

Following \citet{mallen2023not}, we define our first predictor---question subject popularity (\texttt{SPop})---as the mean number of monthly views of the subject entity's Wikipedia page. In the PopQA datset, the \texttt{SPop} score varies from 2 to over 15M.

\subsubsection{Semantic Consistency (\texttt{SCons})}
\label{sec:semantic-consistency-predictor}

Semantic consistency---as defined by the \texttt{SCons} metric in Section~\ref{ssec:consistency-metrics}---associated with $q$, is measured as $\texttt{SCons}(q){=}\textrm{int\_sim}(\mathcal{A})$, where $\mathcal{A}$ consists of greedily-generated answers to $q$ itself and the set of its paraphrases $P(q)$.
%

\subsubsection{Answer Certainty (\texttt{Cert})}
\label{sec:semantic-certainty}

\begin{table*}[h!]
\centering
\resizebox{\textwidth}{!}{
\begin{tabular}{l|l|c}
question & sampled answers & certainty score \\ \hline
What is Robby Krieger’s occupation? & (guitarist, guitarist, guitarist, guitarist, guitarist) & 1.000 \\
What is Shozaburo Nakamura's occupation? & (samurai, samurai, film director, actor, director) & 0.250 \\
What is the capital of Benin? & (cotonou, bamako, abidjan, bamako, bamako ) & 0.521 \\

\end{tabular}
}
\caption{Examples of \texttt{Cert} score assigned to a set of sampled answers to the same question. Notably, cultural bias(es) in contemporary LLMs are manifested by the ``samurai'' answer to the question about Japanese politician.}
\label{tbl:certainty-scores}
\end{table*}

Multiple studies investigated the \textit{uncertainty} of natural language generation in the context of free-form QA. \citet{kuhn2022semantic} put forward a hypothesis that given some degree of freedom (i.e., sampling, not greedy generation), ``$\dots$very \textit{uncertain} generations should be less likely to be \textit{correct}''. Specifically, the authors suggest that a (non-greedily-probed) model producing multiple distinct answers for the same question is unstable and less robust, potentially affecting the model's ability to provide a correct answer to the question.

Uncertainty of a set of answers $\mathcal{A}$ to a factual question $q$ is manifested by the relative amount of distinct answers out of the entire answer pool $\mathcal{A}$. Multiple metrics were suggested to measure uncertainty---or, its complementary metric, \textit{certainty}---of a set of answers, including lexical similarity, Rouge-L \cite{lin2004rouge}, and predictive entropy \citep{kuhn2022semantic}. As with semantic consistency (see Section~\ref{sec:semantic-consistency-predictor}), we found mean pairwise semantic similarity of answers in $\mathcal{A}$ to be the most appropriate metric for certainty of very short factoid answers.  Our sampled answers certainty metric is defined as $\texttt{Cert}(q){=}\textrm{int\_sim}(\mathcal{A})$, where, following \citet{kuhn2022semantic}, $\mathcal{A}$ is a set of ten answers to $q$ sampled non-greedily, setting models' temperature to 0.5. Table~\ref{tbl:certainty-scores} presents several results of sampling answers to questions in the PopQA dataset, along with their respective certainty score. 


\subsubsection{Question Category (\texttt{QCat})}
\label{ssec:question_category}
Figure~\ref{fig:models-correctness} suggests that question category---the semantic grouping a question belongs to---has a considerable effect on an LLM's ability to answer a question correctly. While models systematically succeed in answering questions on \textit{capital}, \textit{color}, and \textit{sport}, they struggle in categories like \textit{director}, \textit{producer}, and \textit{author}. Question category (\texttt{QCat}) has been shown to interact with numerical variables (see Section~\ref{ssec:regression-model}), suggestive of the potential benefits of including question category as a (nominal) categorical variable in our regression analysis.

\subsection{Predictive Model}
\label{ssec:regression-model}
We build a logistic regression model for predicting if an LLM will answer a question correctly. Specifically, for an original question $q{\in}\textrm{PopQA}$, we define a model using the four predictors described in Section~\ref{ssec:preictors}, where the regression outcome is a binary indicator: will $q$ be answered accurately (1), or not (0).\footnote{At inference time, the likelihood of an LLM to provide an accurate answer (i.e., probability value in the 0-1 range) can be considered, instead of the binary target.} We denote the regression response variable by \texttt{correct}, and use \texttt{QCat}, \texttt{SCons}, \texttt{Cert} and \texttt{SPop} as regressors. We apply a natural log transformation to \texttt{SPop}, reducing its skewness, and strengthening its relationship with the target variable.

The regression model assumes a linear relationship between each regressor and the logit of the binary target, holding other regressors constant. We consider the first-order effects of \texttt{QCat} and the numeric variables (\texttt{SCons}, \texttt{Cert} and \texttt{SPop}), as well as the second-order interaction between each numeric variable and question category \texttt{QCat}, where the intuition is that the precise impact of a numeric predictor varies by category.  Figure~\ref{fig:logit_correctness_vs_single_vars} in Appendix~\ref{ssec:logistic_plots} illustrates the need to account for \texttt{QCat} interactions with the numeric regressors because the marginal effect (slope of linear relationship) of each variable on correctness differs by the \texttt{QCat} group.  Consequently, we define our regression model using the common regression notation as
%
\noindent \texttt{correct} {$\mathtt{\sim}$} \texttt{QCat*log(SPop) + QCat*SCons + QCat*Cert},
where `*' denotes the second- and first-order effects of two variables. \texttt{QCat} is treated as a \textit{fixed} rather than \textit{random} categorical effect, since we are interested in the individual effect of each category and do not assume that the relationship types were randomly sampled from the population of available ones.  Appendix~\ref{ssec:regression_tables} table~\ref{tbl:logistic_regression_wald} quantifies the relative contributions of each regressor to the model's goodness of fit, and shows that \texttt{QCat} and its interactions are strongly statistically significant.

Logistic regression is implemented using Python's \texttt{statsmodels} \citep{seabold2010statsmodels} module formula interface. We report regression results when applied on each of the five LLMs detailed in Section~\ref{ssec:models}, further explore the relative contribution of each predictor, and perform an ablation study in the next section.

\subsection{Experimental Results}
\label{ssec:performance-prediction-results}

\begin{table*}[h!]
\centering
\resizebox{\textwidth}{!}{
\begin{tabular}{l|c|c|c||c|c|c}
model & \multicolumn{3}{c||}{\{q $\in$ PopQA\}} & \multicolumn{3}{c}{\{q $\in$ PopQA | correct(\texttt{QCat}(q)) > 0.1\}} \\ \hline
& $\mathrm{R^2}$ & ACC(test set) & mjr. baseline & $\mathrm{R^2}$ & ACC(test set) & mjr. baseline \\ \hline
MT0-XXL (ED) & 0.489 &  0.936 (+2.63) & 0.912 & 0.308 & 0.809 (+18.27) & 0.684 \\
Flan-UL2 (ED) & 0.479 & 0.915 (+4.69) & 0.874 & 0.344 & 0.829 (+19.10) & 0.696 \\
Flan-T5-XXL (ED) & 0.491 & 0.928 (+3.80) & 0.894 & 0.311 & 0.794 (+26.23) & 0.629 \\ \hline
MPT-Instruct2 (D) & 0.430 & 0.878 (+13.1) & 0.776 & 0.425 & 0.862 (+13.40) & 0.760 \\
GPT-NeoX (D) & 0.418 & 0.883 (+8.21) & 0.816 & 0.310 & 0.791 (+22.82) & 0.644 \\
\end{tabular}
}
\caption{QA performance prediction using logistic regression with various models. `ED' stands for encoder-decoder models, `D' -- for decoder-only. McFadden's pseudo R-squared is reported as well as models' accuracy on held-out test set (20\%); relative performance improvement, compared to the baseline, is specified with `+' in parenthesis. Left: all question categories are considered, right: only categories with correctness exceeding 0.1 are considered.}
\label{tbl:models-regression-results}
\end{table*}

\begin{table*}[h!]
\centering
\resizebox{\textwidth}{!}{
\begin{tabular}{l|c|c||c|c}
included predictors & \multicolumn{2}{c||}{\{q $\in$ PopQA\}} & \multicolumn{2}{c}{\{q $\in$ PopQA | correct(\texttt{QCat}(q)) > 0.1\}} \\ \hline
& $\mathrm{R^2}$ & ACC(test set) & $\mathrm{R^2}$ & ACC(test set) \\ \hline
\texttt{SPop}, \texttt{QCat}, \texttt{SCons}, \texttt{Cert} (the full model) & 0.479 & \textbf{0.915 (+4.69)} & 0.344 & \textbf{0.829 (+19.10)} \\
\texttt{QCat}, \texttt{SCons}, \texttt{Cert} & 0.442 & 0.911 (+4.23) & 0.298 & 0.803 (+15.37) \\
\texttt{SPop}, \texttt{SCons}, \texttt{Cert} & 0.362 & 0.904 (+3.43) & 0.207 & 0.782 (+12.35) \\
\texttt{SCons}, \texttt{Cert} & 0.352 & 0.902 (+3.20) & 0.196 & 0.781 (+12.21) \\
\texttt{Cert} & 0.328 & 0.892 (+2.06) & 0.156 & 0.769 (+10.48) \\
\texttt{SCons} & 0.264 & 0.886 (+1.37) & 0.164 & 0.753 (+08.18) \\

\end{tabular}
}
\caption{Ablation analysis with one of the best performing models (Flan-UL2), testing various predictor combinations. The majority vote baseline of Flan-UL2 is 0.874 for the full set of questions, and 0.696 for questions in categories with baseline correctness>0.1. High accuracy, in particular, much higher than baseline, is maintained when omitting \texttt{QCat}; omitting both (not easily obtainable) \texttt{QCat} and \texttt{SPop} results in yet powerful regression model, improving the baseline by 3.20 and 12.21 percent, for the full and selective question set, respectively.}
\label{tbl:regression-ablation}
\end{table*}

\begin{table}[h!]
\centering
\resizebox{\columnwidth}{!}{
\begin{tabular}{l|l|c|c|c|c}
& predictor & $\hat{\beta}$ & [0.025 & 0.975] & p-value \\ \hline
\multirow{3}{*}{\STAB{\rotatebox[origin=c]{90}{{capital}}}}
& \texttt{intercept} & 0.83 & 0.67 & 1.04 & 0.114 \\     
& \texttt{log(SPop)} & 1.64 & 1.28 & 2.10 & 0.000 \\
& \texttt{SCons} & 2.68 & 1.98 & 3.63 & 0.000\\
& \texttt{Cert} & 2.29 & 1.68 & 3.12 & 0.000 \\ \hline \hline
\multirow{3}{*}{\STAB{\rotatebox[origin=c]{90}{{sport}}}}
& \texttt{intercept} & 1.35 & 1.12 & 1.63 & 0.001 \\     
& \texttt{log(SPop)} & 0.94 & 0.77 & 1.15 & 0.584 \\
& \texttt{SCons} & 1.82 & 1.40 & 2.37 & 0.000 \\
& \texttt{Cert} & 1.47 & 1.15 & 1.88 & 0.002 \\
\end{tabular}
}
\caption{
Logistic regression summary of the Flan-UL2 model for two of its best-performing categories: \textit{capital} and \textit{sport}. Variable are standardized (to have a mean of 0.0 and STD of 1.0) for comparative analysis of coefficients. Appendix~\ref{ssec:regression_tables} (tables~\ref{tbl:logistic_regression_flan_xxl}--\ref{tbl:logistic_regression_gpt_neox}) reports full regression models' results, including variable interactions, for all LLMs in this study.}
\label{tbl:regression-summary}
\end{table}

\paragraph{Main Results}
Table~\ref{tbl:models-regression-results} reports the performance of the logistic regression trained on each of the five LLMs. A regression model's goodness of fit is measured using McFadden's pseudo-$\mathrm{R^2}$; according to \citet{mcfadden977}, values of 0.2 and above indicate very good fit.\footnote{Note that this measure does not adjust for the number of regression terms. In appendices we also report AIC, which penalizes models with excessive number of regressors.}  We also report regression models' accuracy on the 20\% held-out test set, where the accuracy should be interpreted in terms of the relative percent increase, compared to the majority vote baseline -- fixing all predictions to 0, due to the higher prior of incorrect answers for questions in PopQA, for all five LLMs in this study. Notably, the random choice baseline is 0.5. 

We repeat the experiment in a more balanced (and desirable) setting, where the set of question categories is limited to those an LLM shows over 10\% correctness on. Naturally, the lower (but still negative) prior, is reflected in the lower majority voice baseline, posing higher prediction difficulty for the regression model. We show (Table~\ref{tbl:models-regression-results}, right) that the benefits of the suggested approach are amplified in this setting: models obtain high accuracy, improving over the majority vote baseline by a significant extent, between 13.40--26.23\%.

\paragraph{Ablation Study}
Next we test the robustness of the regression model, by eliminating regressors, one by one, from an example LLM regression model, and inspecting the outcome, as reported in Table~\ref{tbl:regression-ablation}. Again, we perform this experiment with all question categories, and the set of categories with the correctness prior > 0.1, for the selected model. High prediction accuracy (0.902 and 0.781) is maintained, even when removing \texttt{SPop} and \texttt{QCat}, thereby only including regressors independent of external knowledge---semantic consistency and certainty---predictors that can be computed automatically (including paraphrase generation). Moreover, using only semantic consistency or certainty as a single predictor shows considerable performance gains, in both settings.

\paragraph{In-category Coefficient Analysis}
The ablation study findings are further supported by the regression summary in Table~\ref{tbl:regression-summary}, for two sample question categories with high correctness: \textit{capital} and \textit{sport}. Regressor coefficients ($\hat{\beta}$), as well as their 95\% confidence intervals, and p-values are presented. Positive coefficients reflect the (expected) positive correlation between the predictors and the regression model outcome: higher semantic consistency, higher certainty or question subject popularity are predictive of higher LLM's answer accuracy with respect to the question at hand.

\section{Related Work}
\label{sec:rel-work}

\paragraph{Semantic Consistency of LLMs}
Studies in the domain of model consistency were pioneered with the work by \citet{elazar2021measuring}, who investigated this question in the context of masked language models, where the same factual knowledge (in the form of a single token) was masked from multiple meaning-preserving alternations of the same statement. \citet{fierro2022factual} extended the factual consistency study on MLMs to the multilingual setup. \citet{jang2022becel} extend the notion of consistency to six \textit{behavioral} consistency properties, including semantic textual similarity, machine-reading comprehension, and topic classification. The authors make use of adapted and newly-created datasets for testing multiple fine-tuned language models on the set of selected tasks. Factual consistency experiments are explicitly excluded from the set of tests. Multiple semantic consistency metrics were evaluated by \citet{raj2022measuring} on automatically generated paraphrases of (mostly not factoid) open-domain questions in the TruthfulQA dataset \citep{lin2022truthfulqa}; the authors demonstrate that NLI-based consistency metric correlates best with human judgements, when evaluating the consistency of sentence-length answers. 

A wider notion of \textit{prompt consistency} was studied by \citet{zhou2022prompt} for multiple tasks: NLI, co-reference resolution, word sense disambiguation and sentence completion. The authors design pairwise distillation loss that encourages consistency between semantically-equivalent pair of prompts, and demonstrate increase of over 10\% in models' performance. Finally, \citet{newman2021p} introduce P-Adapters for increasing the robustness of MLMs (specifically, BERT \citep{devlin2018bert}) to prompt alternations. No prior work, to the best of our knowledge, has explicitly addressed the task of LLMs factual semantic consistency, with a high-quality benchmark factual QA dataset.

\paragraph{QA Performance Prediction}
Inspired by the established and well-studied task of \textit{query performance prediction} (QPP) in the domain of information retrieval (i.e., search engines), we develop a framework for predicting the correctness of a generative (not retrieval-based) LLM's response to a factual question -- \textit{question answering performance prediction}. Given a question, the ultimate goal is to score the likelihood of the model to answer the question correctly, without any reference answers. The open-domain nature of questions pose a special challenge for the task, in the complete absence of information facilitating reference-less evaluation, such as a document for the task of summarization, or a paragraph for context-based extractive QA.

Despite its evident importance, prior work on QA performance prediction is relatively scarce. \citet{kuhn2022semantic} have shown that semantic certainty---the consistency of a model's answers to a question, where sampling is allowed---is indicative of the model's ability to answer the question correctly. Specifically, they report that "... [when sampling is allowed] Incorrectly answered questions have more semantically distinct answers than correct ones." Introducing the PopQA dataset of factual questions, \citet{mallen2023not} suggest that factual knowledge memorization depends on the \textit{popularity} of the entity, the subject of a question refers to: the frequency of information about the question subject on the web.
\section{Conclusions}
\label{sec:conclusions}

We explore the robustness of LLMs to paraphrases in the context of open-domain zero-shot QA. Introducing a large and carefully-curated extension of the PopQA dataset (PopQA-TP), with high-quality paraphrases, we first benchmark the semantic consistency of diverse LLMs; next, we develop a framework for QA performance prediction, incorporating semantic consistency, as well as additional aspects, shown to correlate with model's QA accuracy. Collectively, our work shows that a model's ability to answer a question accurately can be reliably predicted, in a reference-less setting.
Our future work includes the exploration of how the semantic consistency metric used in this work can be adapted to additional generative tasks with long(er) answers, e.g., summarization, dialogue.

\section{Limitations}
\label{sec:limitations}

Our study has several limitations: First, the semantic consistency measurement has been studied in the relatively narrow context of the factual QA task; it would be useful to explore how this metric applies and should possible be adapter for additional generative tasks, such as summarization, translation, or QA with free-form long(er) answers. Second, the presented QA performance prediction framework exhibits best results with the full set of predictors, exploiting ``external knowledge''--- subject popularity and question category; those are not always available. Given said that, we show significant prediction benefits even when using easily-obtainable predictors, \texttt{Scons} and \texttt{Cert}.


\bibliographystyle{acl_natbib}
\bibliography{main}

\begin{thebibliography}{26}
\expandafter\ifx\csname natexlab\endcsname\relax\def\natexlab#1{#1}\fi

\bibitem[{Bandel et~al.(2022)Bandel, Aharonov, Shmueli-Scheuer, Shnayderman,
  Slonim, and Ein-Dor}]{bandel2002quality}
Elron Bandel, Ranit Aharonov, Michal Shmueli-Scheuer, Ilya Shnayderman, Noam
  Slonim, and Liat Ein-Dor. 2022.
\newblock \href {https://doi.org/10.18653/v1/2022.acl-long.45} {Quality
  controlled paraphrase generation}.
\newblock In \emph{Proceedings of the 60th Annual Meeting of the Association
  for Computational Linguistics (Volume 1: Long Papers)}, pages 596--609,
  Dublin, Ireland. Association for Computational Linguistics.

\bibitem[{Black et~al.(2022)Black, Biderman, Hallahan, Anthony, Gao, Golding,
  He, Leahy, McDonell, Phang, Pieler, Prashanth, Purohit, Reynolds, Tow, Wang,
  and Weinbach}]{eleutherai_gpt_neox}
Sid Black, Stella Biderman, Eric Hallahan, Quentin Anthony, Leo Gao, Laurence
  Golding, Horace He, Connor Leahy, Kyle McDonell, Jason Phang, Michael Pieler,
  USVSN~Sai Prashanth, Shivanshu Purohit, Laria Reynolds, Jonathan Tow, Ben
  Wang, and Samuel Weinbach. 2022.
\newblock \href {https://doi.org/10.48550/ARXIV.2204.06745} {{GPT-NeoX-20B}: An
  open-source autoregressive language model}.

\bibitem[{Carmel and Kurland(2012)}]{carmel2012query}
David Carmel and Oren Kurland. 2012.
\newblock \href
  {https://doi.org/https://dl.acm.org/doi/abs/10.1145/2348283.2348540} {Query
  performance prediction for {IR}}.
\newblock In \emph{Proceedings of the 35th international ACM SIGIR conference
  on Research and development in information retrieval}, pages 1196--1197.

\bibitem[{Chung et~al.(2022)Chung, Hou, Longpre, Zoph, Tay, Fedus, Li, Wang,
  Dehghani, Brahma et~al.}]{chung2022scaling}
Hyung~Won Chung, Le~Hou, Shayne Longpre, Barret Zoph, Yi~Tay, William Fedus,
  Eric Li, Xuezhi Wang, Mostafa Dehghani, Siddhartha Brahma, et~al. 2022.
\newblock \href {https://arxiv.org/abs/2210.11416} {Scaling
  instruction-finetuned language models}.
\newblock \emph{arXiv preprint arXiv:2210.11416}.

\bibitem[{Devlin et~al.(2018)Devlin, Chang, Lee, and
  Toutanova}]{devlin2018bert}
Jacob Devlin, Ming-Wei Chang, Kenton Lee, and Kristina Toutanova. 2018.
\newblock \href {https://arxiv.org/abs/1810.04805} {{BERT}: Pre-training of
  deep bidirectional transformers for language understanding}.
\newblock \emph{arXiv preprint arXiv:1810.04805}.

\bibitem[{Elazar et~al.(2021)Elazar, Kassner, Ravfogel, Ravichander, Hovy,
  Sch{\"u}tze, and Goldberg}]{elazar2021measuring}
Yanai Elazar, Nora Kassner, Shauli Ravfogel, Abhilasha Ravichander, Eduard
  Hovy, Hinrich Sch{\"u}tze, and Yoav Goldberg. 2021.
\newblock \href {https://aclanthology.org/2021.tacl-1.60/} {Measuring and
  improving consistency in pretrained language models}.
\newblock \emph{Transactions of the Association for Computational Linguistics},
  9:1012--1031.

\bibitem[{Faggioli et~al.(2023)Faggioli, Formal, Marchesin, Clinchant, Ferro,
  and Piwowarski}]{faggioli2023query}
Guglielmo Faggioli, Thibault Formal, Stefano Marchesin, St{\'e}phane Clinchant,
  Nicola Ferro, and Benjamin Piwowarski. 2023.
\newblock \href {https://doi.org/10.48550/arXiv.2302.09947} {Query performance
  prediction for neural {IR}: Are we there yet?}
\newblock In \emph{European Conference on Information Retrieval}, pages
  232--248. Springer.

\bibitem[{Fierro and S{\o}gaard(2022)}]{fierro2022factual}
Constanza Fierro and Anders S{\o}gaard. 2022.
\newblock \href {https://aclanthology.org/2022.findings-acl.240/} {Factual
  consistency of multilingual pretrained language models}.
\newblock In \emph{Findings of the Association for Computational Linguistics:
  ACL 2022}, pages 3046--3052.

\bibitem[{Jang et~al.(2021)Jang, Kwon, and Lukasiewicz}]{jang2021accurate}
Myeongjun Jang, Deuk~Sin Kwon, and Thomas Lukasiewicz. 2021.
\newblock \href {https://doi.org/10.48550/arXiv.2108.06665} {Accurate, yet
  inconsistent? consistency analysis on language understanding models}.
\newblock \emph{arXiv preprint arXiv:2108.06665}.

\bibitem[{Jang et~al.(2022)Jang, Kwon, and Lukasiewicz}]{jang2022becel}
Myeongjun Jang, Deuk~Sin Kwon, and Thomas Lukasiewicz. 2022.
\newblock \href {https://aclanthology.org/2022.coling-1.324/} {{BECEL:
  Benchmark for Consistency Evaluation of Language Models}}.
\newblock In \emph{Proceedings of the 29th International Conference on
  Computational Linguistics}, pages 3680--3696.

\bibitem[{Kuhn et~al.(2022)Kuhn, Gal, and Farquhar}]{kuhn2022semantic}
Lorenz Kuhn, Yarin Gal, and Sebastian Farquhar. 2022.
\newblock \href {https://doi.org/https://doi.org/10.48550/arXiv.2302.09664}
  {Semantic uncertainty: Linguistic invariances for uncertainty estimation in
  natural language generation}.
\newblock In \emph{The Eleventh International Conference on Learning
  Representations}.

\bibitem[{Lin and Och(2004)}]{lin2004rouge}
Chin{-}Yew Lin and Franz~Josef Och. 2004.
\newblock \href {https://doi.org/10.3115/1218955.1219032} {Automatic evaluation
  of machine translation quality using longest common subsequence and
  skip-bigram statistics}.
\newblock In \emph{Proceedings of the 42nd Annual Meeting of the Association
  for Computational Linguistics, 21-26 July, 2004, Barcelona, Spain}, pages
  605--612. {ACL}.

\bibitem[{Lin et~al.(2022)Lin, Hilton, and Evans}]{lin2022truthfulqa}
Stephanie Lin, Jacob Hilton, and Owain Evans. 2022.
\newblock \href {https://aclanthology.org/2022.acl-long.229/} {Truthful{QA}:
  Measuring how models mimic human falsehoods}.
\newblock In \emph{Proceedings of the 60th Annual Meeting of the Association
  for Computational Linguistics (Volume 1: Long Papers)}, pages 3214--3252.

\bibitem[{Mallen et~al.(2023)Mallen, Asai, Zhong, Das, Khashabi, and
  Hajishirzi}]{mallen2023not}
Alex Mallen, Akari Asai, Victor Zhong, Rajarshi Das, Daniel Khashabi, and
  Hannaneh Hajishirzi. 2023.
\newblock \href {https://aclanthology.org/2023.acl-long.546/} {When not to
  trust language models: Investigating effectiveness of parametric and
  non-parametric memories}.
\newblock In \emph{Proceedings of the 61st Annual Meeting of the Association
  for Computational Linguistics (Volume 1: Long Papers)}, pages 9802--9822.

\bibitem[{McFadden(1977)}]{mcfadden977}
Daniel McFadden. 1977.
\newblock \href
  {https://elischolar.library.yale.edu/cowles-discussion-paper-series/707/}
  {Quantitative methods for analyzing travel behaviour of individuals: Some
  recent developments}.
\newblock Cowles Foundation Discussion Papers 474, Cowles Foundation for
  Research in Economics, Yale University.

\bibitem[{MosaicML(2023)}]{mosaicml_mpt}
NLP~Team MosaicML. 2023.
\newblock \href {https://www.mosaicml.com/blog/mpt-7b} {Introducing {MPT-7B}: A
  new standard for open-source, commercially usable {LLM}s}.
\newblock Accessed: 2023-08-05.

\bibitem[{Muennighoff et~al.(2022)Muennighoff, Wang, Sutawika, Roberts,
  Biderman, Scao, Bari, Shen, Yong, Schoelkopf et~al.}]{bigscience_mt0_xxl}
Niklas Muennighoff, Thomas Wang, Lintang Sutawika, Adam Roberts, Stella
  Biderman, Teven~Le Scao, M~Saiful Bari, Sheng Shen, Zheng-Xin Yong, Hailey
  Schoelkopf, et~al. 2022.
\newblock \href {https://aclanthology.org/2023.acl-long.891/} {Crosslingual
  generalization through multitask finetuning}.
\newblock \emph{arXiv preprint arXiv:2211.01786}.

\bibitem[{Newman et~al.(2021)Newman, Choubey, and Rajani}]{newman2021p}
Benjamin Newman, Prafulla~Kumar Choubey, and Nazneen Rajani. 2021.
\newblock \href {https://doi.org/10.48550/arXiv.2110.07280} {P-adapters:
  Robustly extracting factual information from language models with diverse
  prompts}.
\newblock In \emph{International Conference on Learning Representations}.

\bibitem[{Rahamim et~al.(2023)Rahamim, Uziel, Goldbraich, and
  Tavor}]{rahamim2023text}
Adir Rahamim, Guy Uziel, Esther Goldbraich, and Ateret~Anaby Tavor. 2023.
\newblock \href {https://aclanthology.org/2023.findings-acl.466/} {Text
  augmentation using dataset reconstruction for low-resource classification}.
\newblock In \emph{Findings of the Association for Computational Linguistics:
  ACL 2023}, pages 7389--7402.

\bibitem[{Raiber and Kurland(2014)}]{raiber2014query}
Fiana Raiber and Oren Kurland. 2014.
\newblock \href
  {https://doi.org/https://dl.acm.org/doi/10.1145/2600428.2609581}
  {Query-performance prediction: setting the expectations straight}.
\newblock In \emph{Proceedings of the 37th international ACM SIGIR conference
  on Research \& development in information retrieval}, pages 13--22.

\bibitem[{Raj et~al.(2022)Raj, Rosati, and Majumdar}]{raj2022measuring}
Harsh Raj, Domenic Rosati, and Subhabrata Majumdar. 2022.
\newblock \href {https://doi.org/https://doi.org/10.48550/arXiv.2211.05853}
  {Measuring reliability of large language models through semantic
  consistency}.
\newblock In \emph{NeurIPS ML Safety Workshop}.

\bibitem[{Reimers and Gurevych(2019)}]{reimers2019sentence}
Nils Reimers and Iryna Gurevych. 2019.
\newblock \href {https://aclanthology.org/D19-1410/} {{S}entence-{BERT}:
  {S}entence {E}mbeddings using {S}iamese {BERT}-{N}etworks}.
\newblock In \emph{Proceedings of the 2019 Conference on Empirical Methods in
  Natural Language Processing and the 9th International Joint Conference on
  Natural Language Processing (EMNLP-IJCNLP)}, pages 3982--3992.

\bibitem[{Seabold and Perktold(2010)}]{seabold2010statsmodels}
Skipper Seabold and Josef Perktold. 2010.
\newblock \href {https://doi.org/10.25080/Majora-92bf1922-011} {statsmodels:
  Econometric and statistical modeling with python}.
\newblock In \emph{9th Python in Science Conference}.

\bibitem[{Tay(2023)}]{flan_ul2}
Yi~Tay. 2023.
\newblock \href {https://www.yitay.net/blog/flan-ul2-20b} {{A new open source
  Flan 20B with UL2}}.

\bibitem[{Zhou et~al.(2022)Zhou, He, Ma, Berg-Kirkpatrick, and
  Neubig}]{zhou2022prompt}
Chunting Zhou, Junxian He, Xuezhe Ma, Taylor Berg-Kirkpatrick, and Graham
  Neubig. 2022.
\newblock \href {https://aclanthology.org/2022.findings-emnlp.192/} {Prompt
  consistency for zero-shot task generalization}.
\newblock In \emph{Findings of the Association for Computational Linguistics:
  EMNLP 2022}, pages 2613--2626.

\bibitem[{Zhou and Croft(2007)}]{zhou2007query}
Yun Zhou and W~Bruce Croft. 2007.
\newblock \href {https://dl.acm.org/doi/10.1145/1277741.1277835} {Query
  performance prediction in web search environments}.
\newblock In \emph{Proceedings of the 30th annual international ACM SIGIR
  conference on Research and development in information retrieval}, pages
  543--550.

\end{thebibliography}

\appendix
\section{Appendices}

\subsection{Logistic Regression Diagnostic Plots
\label{ssec:logistic_plots}
}

As mentioned in Section~\ref{ssec:regression-model}, in logistic regression, a binary response $y$ (in our case, the indicator $\texttt{correct}{\in}\{0,1\}$) is modeled as a function of a set of regressors; the regressors consist of certain predictor variables and possible  interactions between them.  More precisely, the logit transformation of the dependent variable $p=\textrm{Pr}(y=1)$, the probability of the indicator equaling 1 (denoted \texttt{pcorrect}) is modeled as a linear function of the regressors; thus, the logit should have a linear relationship with each regressor.

\paragraph{Identifying Predictor Interactions} Our chosen logistic model is \texttt{correct} {$\mathtt{\sim}$} \texttt{QCat*log(SPop) + QCat*SCons + QCat*Cert}.  The appropriateness of the addition of a regressor in the logistic model can be visually analyzed by plotting the empirical values of $p$ (here, \texttt{pcorrect}) conditioned on values of a regressor. Here, we we illustrate with the interaction of the categorical \texttt{QCat} with each numeric variable $x {\in} \{\texttt{log(SPop)}, \texttt{SCons}, \texttt{Cert}\}$. The interaction means that the slope of the estimated linear relationship between \texttt{pcorrect} and each variable $x$ can differ conditionally on each level of the categorical \texttt{QCat}.  If the interaction is significant, we should see significant slope differences for at least some of the levels of \texttt{QCat}; if there is no interaction, the lines will have similar slope but possibly differing vertical displacement (i.e., vertical intercepts). 

\begin{figure*}[h!]
\centering
\includegraphics[width=\textwidth]{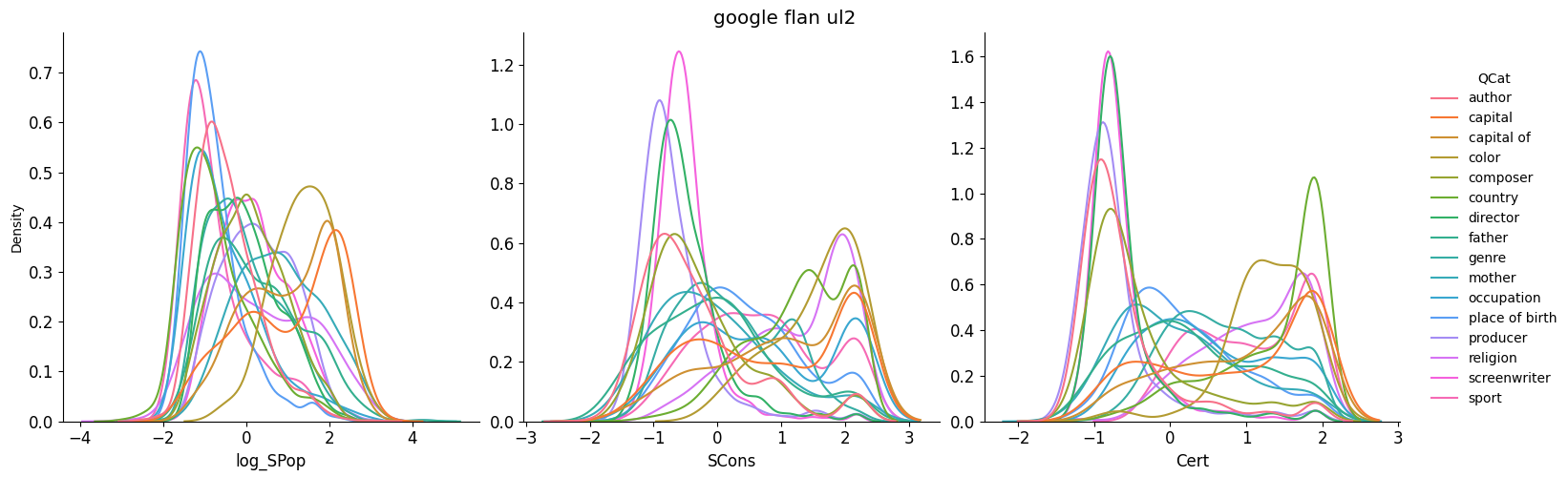}
\caption{Kernel density plots for numeric variables, conditional on each level of \texttt{QCat}.  These are fixed in PopQA and thus do not depend on the language model.
\label{fig:single_vars_kdes}}
\end{figure*}

\begin{figure*}[h!]
\centering
\includegraphics[width=\textwidth]{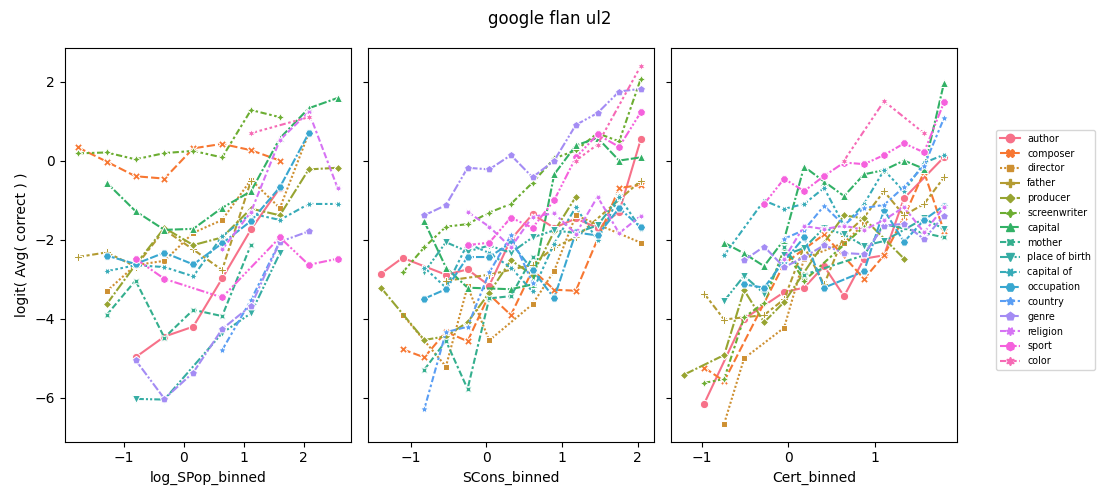}
\caption{Line plots of average value of observed \texttt{correct}, conditional on levels of \texttt{QCat}, for each numeric variable within a given interval of its range.
\label{fig:logit_correctness_vs_single_vars}}
\end{figure*}

Because the continuous-valued \texttt{pcorrect} is not observed (we see only the binary \texttt{correct}), we can approximate it by first, binning the observed range of each variable into, say, 15 equal-width bins; second, restricting to observations with value of $x$ in a given bin and a given value of \texttt{QCat}, and calculating the average value of \texttt{correct} for these, we can approximate the typical value of \texttt{pcorrect} (assuming that there are enough observations in the subset) for $x$ in that bin interval. In Figure~\ref{fig:logit_correctness_vs_single_vars}, we plot this estimated value of \texttt{pcorrect} versus the bin midpoint, considering only bins of $x$ falling between in the center 95\% interval of observed $x$ values for that level of \texttt{QCat} (see Figure~\ref{fig:single_vars_kdes}), to reduce noisy estimates at the edges.  

Figure~\ref{fig:logit_correctness_vs_single_vars} shows that the presence of an interaction is reasonable, since for each variable, the relationship is roughly linear for each value of \texttt{QCat} but that the slopes often differ; the differing vertical displacements of the lines for each variable $x$ are modeled by the single-order coefficients of \texttt{QCat}.

\subsection{Logistic Regression Coefficient Tables
\label{ssec:regression_tables}
}

Here we present summary tables from the logistic model in Section~\ref{ssec:logistic_plots} fit to the results of each LLM on PopQA-TP, without a train-test split.  
Table~\ref{tbl:logistic_regression_fits} summarizes the overall fit of the chosen logistic model on each LLM. The McFadden's statistic measures overall goodness-of-fit without penalizing the number of regressors; since statistic values over 0.4 indicate excellent fit, the model fits very well for each LLM.  The Akaike Information Criterion (AIC) statistic adjusts for the number of regressors, and this model specification achieved the lowest (best) AIC for each LLM over the reduced models, indicating that the interaction effects are correctly included in the predictive logistic model.

\begin{table}[hbt]
\centering
\begin{tabular}{l|rr}
model & McFadden's $\mathrm{R^2}$ & AIC \\
\hline
Flan-T5-XXL & 0.492 & 5029.942 \\
Flan-UL2 & 0.479 & 5746.113 \\
MT0-XXL & 0.490 & 4471.597 \\
MPT-Instruct2 & 0.430 & 8770.356 \\
GPT-NeoX & 0.419 & 8040.402 \\
\end{tabular}
\caption{Summary of logistic regression fits by model.
\label{tbl:logistic_regression_fits}}
\end{table}

\begin{table*}[h!]
\centering
\resizebox{\textwidth}{!}{
\begin{tabular}{lr|rr|rr|rr|rr|rr}
\multicolumn{2}{c|}{regressor} &
\multicolumn{2}{c|}{Flan-T5-XXL} &
\multicolumn{2}{c|}{Flan-UL2} &
\multicolumn{2}{c|}{MT0-XXL} &
\multicolumn{2}{c|}{MPT-Intruct2} &
\multicolumn{2}{c}{GPT-NeoX-20B}\\
\hline
mame & df & stat. & symbol & stat. & symbol & stat. & symbol & stat. & symbol & stat. & symbol \\
\hline
intercept & 1 & 115.45 & *** & 184.73 & *** & 42.26 & *** & 220.42 & *** & 351.11 & *** \\
QCat & 15 & 172.68 & *** & 245.25 & *** & 175.08 & *** & 499.15 & *** & 873.53 & *** \\
log\_SPop & 1 & 16.08 & *** & 22.78 & *** & 8.76 & ** & 165.84 & *** & 172.14 & *** \\
SCons & 1 & 0.10 &  & 1.38 &  & 6.29 & * & 4.51 & * & 2.35 &  \\
Cert & 1 & 34.87 & *** & 34.39 & *** & 13.06 & *** & 76.95 & *** & 15.09 & *** \\
\hline
QCat:log\_SPop & 15 & 88.27 & *** & 126.21 & *** & 78.63 & *** & 420.01 & *** & 393.47 & *** \\
QCat:SCons & 15 & 29.15 & * & 32.45 & ** & 22.47 & . & 65.27 & *** & 69.93 & *** \\
QCat:Cert & 15 & 92.00 & *** & 71.28 & *** & 82.93 & *** & 116.04 & *** & 61.94 & *** \\
\end{tabular}
}
\caption{Logistic regression Wald statistics for each language model.
\label{tbl:logistic_regression_wald}}
\end{table*}

Table~\ref{tbl:logistic_regression_wald} quantifies how much each regressor in the logistic model contributes to the overall fit of the model. This can be assessed by comparing the magnitudes of the Wald $\chi^2$ statistics (``stat.'') for different regressors in the same LLM, and across different LLMs.  The statistical significance of each is indicated by the ``symbol'' column, which codes\footnote{This notational convention is used in \texttt{R} statistical software.} the statistic's p-value: *** ($<0.001$), ** ($<0.01$), * ($<0.05$), . ($<0.1$), or blank ($\geq 0.1$).  The statistical significance penalizes the regressor's constraint degrees of freedom (`df' column), which equals 1 for numeric variables and \#levels-1 for a categorical variable; hence here the numeric interactions with \texttt{QCat} have 15 degrees of freedom, since there are 16 categories.

Overall, the question category \texttt{QCat} has the most explanatory power (its statistic is the largest), followed by \texttt{Cert} or \texttt{log(SPop)}; \texttt{SCons} contributes relatively little on its own, but more when it is interacted with \texttt{QCat}.  Interestingly, the contributions of \texttt{log(SPop)}, \texttt{Cert}, the \texttt{log(SPop)}-\texttt{QCat} interactions are much larger in the encoder-only LLMs (MPT-Instruct2 and GPT-NeoX-20B) compared to the encoder-decoder language models, though the interactions in both cases are already very significant (scoring *** regardless).

Though Table~\ref{tbl:logistic_regression_wald} summarizes each regressor's contribution, it does not tell us about the \textit{direction} of the effect of each regressor.  For that, we refer to Tables~\ref{tbl:logistic_regression_flan_xxl}--\ref{tbl:logistic_regression_gpt_neox}, which show the full set of coefficient estimates.  In each table, we have the coefficient estimate ($\hat{\beta}$), its 95\% confidence interval, the p-value, and the symbol coding of the p-value.  The interpretation of a coefficient is the marginal effect on logit(\texttt{correct}) of a 1-unit increase in each regressor.  For the numeric variables, which have been standardized, this corresponds to a 1 standard deviation change (allowing their effect to be compared despite the different original scales); for the factor \texttt{QCat}, this corresponds to the increase in the logit associated with the given category value relative to that of the omitted level, ``author''.  Thus, positive values of the coefficient indicate that regressor, all others being equal, is associated with a positive increase in correctness.  
For \texttt{QCat}, for instance, since the choice of omitted level is arbitrary (it is alphabetical), the coefficient sign only has a relative, not absolute interpretation.  For instance, if the coefficient on \texttt{log(SPop)} is 4.5, and its interaction with \texttt{QCat=color} is $-3.2$, this means that \texttt{log(SPop)} is still positively correlated ($4.1{-}3.2{=}0.9{>}0$) with \texttt{correct} when \texttt{QCat=color}, but that its marginal effect is lower than for ``author''.   Hence, for ease of interpretation, we introduce a ``conditional coefficient'' column, which performs this adjustment to allow each regressor to be evaluated on its own; but it is not the original variable in the regression, hence should only be used in the context of understanding the table values.  
We see that in nearly every value of \texttt{QCat}, the numeric variables \texttt{log(SPop)}, \texttt{SCons}, and \texttt{SCert} have positive values for this column, indicating positive effect on correctness.  This accords with Figure~\ref{fig:logit_correctness_vs_single_vars}, where the lineplots nearly all have positive slopes.

\begin{table*}[hbt]
\small
\centering
\begin{tabular}{l|rrrrlr}
 & coefficient ($\hat{\beta}$)& [0.025 & 0.975] & p-value & symbol & conditional coefficient \\
\hline
intercept & -4.467 & -5.282 & -3.652 & 0.000 & *** & -4.467 \\
QCat[T.capital] & 2.338 & 1.437 & 3.239 & 0.000 & *** & -2.129 \\
QCat[T.capital of] & -0.240 & -1.589 & 1.110 & 0.728 &  & -4.707 \\
QCat[T.color] & -3.205 & -8.514 & 2.104 & 0.237 &  & -7.672 \\
QCat[T.composer] & 0.852 & -0.077 & 1.782 & 0.072 & . & -3.615 \\
QCat[T.country] & 1.844 & 0.921 & 2.767 & 0.000 & *** & -2.623 \\
QCat[T.director] & -4.019 & -7.026 & -1.012 & 0.009 & ** & -8.487 \\
QCat[T.father] & 1.409 & 0.479 & 2.338 & 0.003 & ** & -3.059 \\
QCat[T.genre] & 1.955 & 1.081 & 2.830 & 0.000 & *** & -2.512 \\
QCat[T.mother] & -0.005 & -1.977 & 1.967 & 0.996 &  & -4.473 \\
QCat[T.occupation] & 2.493 & 1.610 & 3.377 & 0.000 & *** & -1.974 \\
QCat[T.place of birth] & 0.480 & -0.655 & 1.614 & 0.407 &  & -3.988 \\
QCat[T.producer] & -1.924 & -4.359 & 0.511 & 0.122 &  & -6.391 \\
QCat[T.religion] & 0.530 & -1.180 & 2.239 & 0.544 &  & -3.938 \\
QCat[T.screenwriter] & -0.124 & -1.456 & 1.209 & 0.856 &  & -4.591 \\
QCat[T.sport] & 3.205 & 2.271 & 4.140 & 0.000 & *** & -1.262 \\
\hline
log\_SPop & 1.125 & 0.575 & 1.674 & 0.000 & *** & 1.125 \\
QCat[T.capital]:log\_SPop & -0.680 & -1.267 & -0.093 & 0.023 & * & 0.444 \\
QCat[T.capital of]:log\_SPop & 0.798 & 0.052 & 1.544 & 0.036 & * & 1.923 \\
QCat[T.color]:log\_SPop & -1.087 & -2.406 & 0.231 & 0.106 &  & 0.037 \\
QCat[T.composer]:log\_SPop & -1.202 & -1.893 & -0.512 & 0.001 & *** & -0.078 \\
QCat[T.country]:log\_SPop & -0.952 & -1.533 & -0.371 & 0.001 & ** & 0.173 \\
QCat[T.director]:log\_SPop & 1.823 & 0.152 & 3.494 & 0.032 & * & 2.948 \\
QCat[T.father]:log\_SPop & -0.789 & -1.418 & -0.160 & 0.014 & * & 0.336 \\
QCat[T.genre]:log\_SPop & -0.766 & -1.348 & -0.184 & 0.010 & ** & 0.358 \\
QCat[T.mother]:log\_SPop & -0.485 & -1.641 & 0.671 & 0.411 &  & 0.640 \\
QCat[T.occupation]:log\_SPop & -0.872 & -1.474 & -0.271 & 0.004 & ** & 0.252 \\
QCat[T.place of birth]:log\_SPop & -1.045 & -1.802 & -0.287 & 0.007 & ** & 0.080 \\
QCat[T.producer]:log\_SPop & 0.528 & -0.883 & 1.940 & 0.463 &  & 1.653 \\
QCat[T.religion]:log\_SPop & -0.301 & -0.921 & 0.320 & 0.342 &  & 0.824 \\
QCat[T.screenwriter]:log\_SPop & -0.881 & -1.883 & 0.121 & 0.085 & . & 0.244 \\
QCat[T.sport]:log\_SPop & -1.118 & -1.716 & -0.519 & 0.000 & *** & 0.007 \\
\hline
SCons & 0.097 & -0.514 & 0.708 & 0.756 &  & 0.097 \\
QCat[T.capital]:SCons & 0.258 & -0.407 & 0.923 & 0.447 &  & 0.355 \\
QCat[T.capital of]:SCons & -0.128 & -0.863 & 0.607 & 0.733 &  & -0.031 \\
QCat[T.color]:SCons & 1.466 & -0.711 & 3.643 & 0.187 &  & 1.563 \\
QCat[T.composer]:SCons & 0.208 & -0.495 & 0.911 & 0.562 &  & 0.305 \\
QCat[T.country]:SCons & 0.904 & 0.210 & 1.598 & 0.011 & * & 1.001 \\
QCat[T.director]:SCons & -0.102 & -1.178 & 0.974 & 0.853 &  & -0.005 \\
QCat[T.father]:SCons & 0.031 & -0.694 & 0.756 & 0.934 &  & 0.128 \\
QCat[T.genre]:SCons & 0.223 & -0.439 & 0.885 & 0.508 &  & 0.320 \\
QCat[T.mother]:SCons & -0.748 & -2.099 & 0.603 & 0.278 &  & -0.651 \\
QCat[T.occupation]:SCons & 0.252 & -0.415 & 0.919 & 0.459 &  & 0.349 \\
QCat[T.place of birth]:SCons & 0.522 & -0.271 & 1.315 & 0.197 &  & 0.619 \\
QCat[T.producer]:SCons & -0.013 & -1.464 & 1.437 & 0.986 &  & 0.084 \\
QCat[T.religion]:SCons & -0.061 & -0.951 & 0.829 & 0.894 &  & 0.036 \\
QCat[T.screenwriter]:SCons & 0.036 & -1.145 & 1.216 & 0.953 &  & 0.133 \\
QCat[T.sport]:SCons & 0.457 & -0.210 & 1.124 & 0.179 &  & 0.554 \\
\hline
Cert & 2.100 & 1.403 & 2.797 & 0.000 & *** & 2.100 \\
QCat[T.capital]:Cert & -0.933 & -1.683 & -0.183 & 0.015 & * & 1.167 \\
QCat[T.capital of]:Cert & -0.869 & -1.741 & 0.003 & 0.051 & . & 1.231 \\
QCat[T.color]:Cert & 1.967 & -0.997 & 4.930 & 0.193 &  & 4.067 \\
QCat[T.composer]:Cert & -1.613 & -2.418 & -0.808 & 0.000 & *** & 0.487 \\
QCat[T.country]:Cert & -1.050 & -1.807 & -0.292 & 0.007 & ** & 1.051 \\
QCat[T.director]:Cert & -1.170 & -2.335 & -0.005 & 0.049 & * & 0.930 \\
QCat[T.father]:Cert & -1.419 & -2.278 & -0.560 & 0.001 & ** & 0.681 \\
QCat[T.genre]:Cert & -2.008 & -2.749 & -1.266 & 0.000 & *** & 0.092 \\
QCat[T.mother]:Cert & -1.922 & -3.959 & 0.116 & 0.065 & . & 0.179 \\
QCat[T.occupation]:Cert & -1.702 & -2.487 & -0.917 & 0.000 & *** & 0.398 \\
QCat[T.place of birth]:Cert & -0.735 & -1.630 & 0.159 & 0.107 &  & 1.365 \\
QCat[T.producer]:Cert & 0.620 & -1.414 & 2.655 & 0.550 &  & 2.720 \\
QCat[T.religion]:Cert & -1.126 & -2.232 & -0.020 & 0.046 & * & 0.974 \\
QCat[T.screenwriter]:Cert & 1.183 & -0.241 & 2.607 & 0.104 &  & 3.283 \\
QCat[T.sport]:Cert & -1.192 & -1.968 & -0.416 & 0.003 & ** & 0.908 \\
\hline
\end{tabular}
\caption{Logistic regression results for model Flan-T5-XXL.
\label{tbl:logistic_regression_flan_xxl}}
\end{table*}

\begin{table*}[hbt]
\small
\centering
\begin{tabular}{l|rrrrlr}
 & coefficient ($\hat{\beta}$)& [0.025 & 0.975] & p-value & symbol & conditional coefficient \\
\hline
intercept & -4.152 & -4.750 & -3.553 & 0.000 & *** & -4.152 \\
QCat[T.capital] & 2.133 & 1.425 & 2.841 & 0.000 & *** & -2.019 \\
QCat[T.capital of] & -0.718 & -1.988 & 0.551 & 0.267 &  & -4.870 \\
QCat[T.color] & 3.100 & -0.626 & 6.826 & 0.103 &  & -1.051 \\
QCat[T.composer] & 0.111 & -0.746 & 0.969 & 0.799 &  & -4.040 \\
QCat[T.country] & 1.129 & 0.328 & 1.931 & 0.006 & ** & -3.023 \\
QCat[T.director] & -1.259 & -2.387 & -0.131 & 0.029 & * & -5.411 \\
QCat[T.father] & 1.102 & 0.334 & 1.870 & 0.005 & ** & -3.050 \\
QCat[T.genre] & 1.642 & 0.986 & 2.297 & 0.000 & *** & -2.510 \\
QCat[T.mother] & 0.030 & -1.572 & 1.632 & 0.971 &  & -4.122 \\
QCat[T.occupation] & 1.633 & 0.871 & 2.394 & 0.000 & *** & -2.519 \\
QCat[T.place of birth] & 1.578 & 0.818 & 2.337 & 0.000 & *** & -2.574 \\
QCat[T.producer] & -0.850 & -2.011 & 0.311 & 0.151 &  & -5.002 \\
QCat[T.religion] & 1.324 & 0.322 & 2.325 & 0.010 & ** & -2.828 \\
QCat[T.screenwriter] & -0.038 & -0.911 & 0.834 & 0.931 &  & -4.190 \\
QCat[T.sport] & 3.386 & 2.671 & 4.101 & 0.000 & *** & -0.766 \\
\hline
log\_SPop & 1.060 & 0.625 & 1.495 & 0.000 & *** & 1.060 \\
QCat[T.capital]:log\_SPop & -0.655 & -1.134 & -0.176 & 0.007 & ** & 0.405 \\
QCat[T.capital of]:log\_SPop & 1.269 & 0.573 & 1.964 & 0.000 & *** & 2.329 \\
QCat[T.color]:log\_SPop & -0.286 & -1.807 & 1.235 & 0.713 &  & 0.774 \\
QCat[T.composer]:log\_SPop & -1.088 & -1.698 & -0.479 & 0.000 & *** & -0.028 \\
QCat[T.country]:log\_SPop & -0.824 & -1.298 & -0.349 & 0.001 & *** & 0.236 \\
QCat[T.director]:log\_SPop & 0.268 & -0.490 & 1.026 & 0.488 &  & 1.328 \\
QCat[T.father]:log\_SPop & -0.783 & -1.295 & -0.270 & 0.003 & ** & 0.277 \\
QCat[T.genre]:log\_SPop & -0.416 & -0.882 & 0.051 & 0.081 & . & 0.644 \\
QCat[T.mother]:log\_SPop & -1.379 & -2.285 & -0.472 & 0.003 & ** & -0.318 \\
QCat[T.occupation]:log\_SPop & -0.459 & -0.972 & 0.054 & 0.079 & . & 0.601 \\
QCat[T.place of birth]:log\_SPop & -0.824 & -1.398 & -0.249 & 0.005 & ** & 0.237 \\
QCat[T.producer]:log\_SPop & 0.339 & -0.440 & 1.119 & 0.394 &  & 1.399 \\
QCat[T.religion]:log\_SPop & -0.148 & -0.665 & 0.369 & 0.575 &  & 0.912 \\
QCat[T.screenwriter]:log\_SPop & -0.040 & -0.678 & 0.597 & 0.901 &  & 1.020 \\
QCat[T.sport]:log\_SPop & -1.126 & -1.620 & -0.631 & 0.000 & *** & -0.065 \\
\hline
SCons & 0.256 & -0.171 & 0.683 & 0.240 &  & 0.256 \\
QCat[T.capital]:SCons & 0.587 & 0.088 & 1.086 & 0.021 & * & 0.843 \\
QCat[T.capital of]:SCons & 0.139 & -0.465 & 0.743 & 0.652 &  & 0.395 \\
QCat[T.color]:SCons & 0.266 & -1.315 & 1.847 & 0.742 &  & 0.522 \\
QCat[T.composer]:SCons & 0.397 & -0.211 & 1.005 & 0.201 &  & 0.653 \\
QCat[T.country]:SCons & 0.553 & 0.034 & 1.071 & 0.037 & * & 0.809 \\
QCat[T.director]:SCons & -0.199 & -0.913 & 0.514 & 0.584 &  & 0.057 \\
QCat[T.father]:SCons & 0.220 & -0.313 & 0.754 & 0.418 &  & 0.477 \\
QCat[T.genre]:SCons & 0.040 & -0.440 & 0.520 & 0.870 &  & 0.296 \\
QCat[T.mother]:SCons & 0.372 & -0.660 & 1.404 & 0.480 &  & 0.628 \\
QCat[T.occupation]:SCons & -0.189 & -0.728 & 0.351 & 0.493 &  & 0.068 \\
QCat[T.place of birth]:SCons & 0.182 & -0.394 & 0.759 & 0.535 &  & 0.438 \\
QCat[T.producer]:SCons & -0.086 & -0.749 & 0.578 & 0.800 &  & 0.171 \\
QCat[T.religion]:SCons & -0.177 & -0.809 & 0.455 & 0.583 &  & 0.079 \\
QCat[T.screenwriter]:SCons & 0.529 & -0.084 & 1.142 & 0.091 & . & 0.785 \\
QCat[T.sport]:SCons & 0.376 & -0.133 & 0.886 & 0.148 &  & 0.633 \\
\hline
Cert & 1.490 & 0.992 & 1.987 & 0.000 & *** & 1.490 \\
QCat[T.capital]:Cert & -0.669 & -1.253 & -0.085 & 0.025 & * & 0.820 \\
QCat[T.capital of]:Cert & -1.007 & -1.703 & -0.312 & 0.004 & ** & 0.482 \\
QCat[T.color]:Cert & -0.833 & -2.490 & 0.824 & 0.324 &  & 0.656 \\
QCat[T.composer]:Cert & -0.123 & -0.828 & 0.582 & 0.732 &  & 1.366 \\
QCat[T.country]:Cert & -0.029 & -0.642 & 0.583 & 0.926 &  & 1.460 \\
QCat[T.director]:Cert & 0.416 & -0.354 & 1.185 & 0.290 &  & 1.905 \\
QCat[T.father]:Cert & -0.439 & -1.086 & 0.209 & 0.184 &  & 1.051 \\
QCat[T.genre]:Cert & -1.177 & -1.736 & -0.617 & 0.000 & *** & 0.313 \\
QCat[T.mother]:Cert & -0.028 & -1.381 & 1.324 & 0.967 &  & 1.461 \\
QCat[T.occupation]:Cert & -0.996 & -1.659 & -0.333 & 0.003 & ** & 0.493 \\
QCat[T.place of birth]:Cert & -1.069 & -1.734 & -0.405 & 0.002 & ** & 0.420 \\
QCat[T.producer]:Cert & -0.438 & -1.134 & 0.259 & 0.218 &  & 1.052 \\
QCat[T.religion]:Cert & -1.045 & -1.813 & -0.278 & 0.008 & ** & 0.444 \\
QCat[T.screenwriter]:Cert & -0.214 & -0.861 & 0.433 & 0.517 &  & 1.275 \\
QCat[T.sport]:Cert & -0.949 & -1.550 & -0.348 & 0.002 & ** & 0.540 \\
\end{tabular}
\caption{Logistic regression results for model Flan-UL2.
\label{tbl:logistic_regression_flan_ul2}}
\end{table*}

\begin{table*}[hbt]
\small
\centering
\begin{tabular}{l|rrrrlr}
 & coefficient ($\hat{\beta}$)& [0.025 & 0.975] & p-value & symbol & conditional coefficient \\
\hline
Intercept & -3.544 & -4.612 & -2.475 & 0.000 & *** & -3.544 \\
QCat[T.capital] & 2.576 & 1.453 & 3.700 & 0.000 & *** & -0.967 \\
QCat[T.capital of] & -0.237 & -1.653 & 1.179 & 0.743 &  & -3.781 \\
QCat[T.color] & 1.247 & -2.831 & 5.325 & 0.549 &  & -2.297 \\
QCat[T.composer] & 2.375 & 0.794 & 3.955 & 0.003 & ** & -1.169 \\
QCat[T.country] & 1.572 & 0.402 & 2.743 & 0.008 & ** & -1.971 \\
QCat[T.director] & -13.903 & -24.854 & -2.952 & 0.013 & * & -17.447 \\
QCat[T.father] & -0.053 & -1.317 & 1.211 & 0.934 &  & -3.597 \\
QCat[T.genre] & 0.054 & -1.133 & 1.242 & 0.929 &  & -3.489 \\
QCat[T.mother] & -0.891 & -3.655 & 1.873 & 0.527 &  & -4.435 \\
QCat[T.occupation] & 1.336 & 0.151 & 2.521 & 0.027 & * & -2.208 \\
QCat[T.place of birth] & -1.173 & -2.588 & 0.243 & 0.104 &  & -4.716 \\
QCat[T.producer] & -5.700 & -9.819 & -1.580 & 0.007 & ** & -9.243 \\
QCat[T.religion] & 0.043 & -1.268 & 1.353 & 0.949 &  & -3.501 \\
QCat[T.screenwriter] & -4.373 & -8.736 & -0.010 & 0.050 & * & -7.916 \\
QCat[T.sport] & 1.063 & -0.098 & 2.223 & 0.073 & . & -2.481 \\
\hline
log\_SPop & 0.907 & 0.307 & 1.508 & 0.003 & ** & 0.907 \\
QCat[T.capital]:log\_SPop & -0.747 & -1.374 & -0.120 & 0.020 & * & 0.161 \\
QCat[T.capital of]:log\_SPop & 0.837 & 0.047 & 1.628 & 0.038 & * & 1.745 \\
QCat[T.color]:log\_SPop & -0.428 & -1.763 & 0.907 & 0.530 &  & 0.480 \\
QCat[T.composer]:log\_SPop & -1.197 & -2.100 & -0.295 & 0.009 & ** & -0.290 \\
QCat[T.country]:log\_SPop & -0.695 & -1.321 & -0.069 & 0.029 & * & 0.212 \\
QCat[T.director]:log\_SPop & 5.929 & 1.201 & 10.657 & 0.014 & * & 6.837 \\
QCat[T.father]:log\_SPop & -0.772 & -1.440 & -0.104 & 0.024 & * & 0.135 \\
QCat[T.genre]:log\_SPop & -0.817 & -1.464 & -0.170 & 0.013 & * & 0.090 \\
QCat[T.mother]:log\_SPop & -0.119 & -1.335 & 1.097 & 0.847 &  & 0.788 \\
QCat[T.occupation]:log\_SPop & -0.390 & -1.043 & 0.262 & 0.241 &  & 0.517 \\
QCat[T.place of birth]:log\_SPop & -0.965 & -1.753 & -0.176 & 0.016 & * & -0.057 \\
QCat[T.producer]:log\_SPop & 2.210 & 0.286 & 4.133 & 0.024 & * & 3.117 \\
QCat[T.religion]:log\_SPop & -0.274 & -0.952 & 0.404 & 0.429 &  & 0.634 \\
QCat[T.screenwriter]:log\_SPop & 0.773 & -0.394 & 1.940 & 0.194 &  & 1.681 \\
QCat[T.sport]:log\_SPop & -0.654 & -1.330 & 0.021 & 0.058 & . & 0.253 \\
\hline
SCons & 0.861 & 0.188 & 1.534 & 0.012 & * & 0.861 \\
QCat[T.capital]:SCons & -0.406 & -1.118 & 0.307 & 0.264 &  & 0.455 \\
QCat[T.capital of]:SCons & -0.641 & -1.402 & 0.121 & 0.099 & . & 0.221 \\
QCat[T.color]:SCons & -1.429 & -4.268 & 1.409 & 0.324 &  & -0.568 \\
QCat[T.composer]:SCons & -0.067 & -0.950 & 0.817 & 0.882 &  & 0.794 \\
QCat[T.country]:SCons & -0.600 & -1.342 & 0.142 & 0.113 &  & 0.262 \\
QCat[T.director]:SCons & 0.582 & -1.432 & 2.596 & 0.571 &  & 1.443 \\
QCat[T.father]:SCons & -1.066 & -1.855 & -0.277 & 0.008 & ** & -0.205 \\
QCat[T.genre]:SCons & -0.354 & -1.111 & 0.402 & 0.359 &  & 0.507 \\
QCat[T.mother]:SCons & 0.709 & -0.902 & 2.319 & 0.388 &  & 1.570 \\
QCat[T.occupation]:SCons & -0.387 & -1.139 & 0.366 & 0.314 &  & 0.474 \\
QCat[T.place of birth]:SCons & -0.684 & -1.497 & 0.128 & 0.099 & . & 0.177 \\
QCat[T.producer]:SCons & -0.489 & -2.023 & 1.045 & 0.532 &  & 0.372 \\
QCat[T.religion]:SCons & 0.086 & -0.855 & 1.027 & 0.858 &  & 0.947 \\
QCat[T.screenwriter]:SCons & -0.655 & -1.729 & 0.419 & 0.232 &  & 0.206 \\
QCat[T.sport]:SCons & -0.380 & -1.157 & 0.397 & 0.338 &  & 0.481 \\
\hline
Cert & 2.119 & 0.970 & 3.268 & 0.000 & *** & 2.119 \\
QCat[T.capital]:Cert & -1.189 & -2.366 & -0.012 & 0.048 & * & 0.930 \\
QCat[T.capital of]:Cert & -1.111 & -2.345 & 0.123 & 0.078 & . & 1.008 \\
QCat[T.color]:Cert & -0.683 & -3.289 & 1.924 & 0.608 &  & 1.437 \\
QCat[T.composer]:Cert & 2.431 & 0.363 & 4.498 & 0.021 & * & 4.550 \\
QCat[T.country]:Cert & -0.976 & -2.167 & 0.215 & 0.108 &  & 1.143 \\
QCat[T.director]:Cert & -1.007 & -3.974 & 1.961 & 0.506 &  & 1.112 \\
QCat[T.father]:Cert & -1.180 & -2.428 & 0.069 & 0.064 & . & 0.939 \\
QCat[T.genre]:Cert & -1.744 & -2.934 & -0.553 & 0.004 & ** & 0.376 \\
QCat[T.mother]:Cert & -0.451 & -2.781 & 1.879 & 0.705 &  & 1.668 \\
QCat[T.occupation]:Cert & -1.850 & -3.101 & -0.599 & 0.004 & ** & 0.269 \\
QCat[T.place of birth]:Cert & -0.615 & -1.854 & 0.625 & 0.331 &  & 1.505 \\
QCat[T.producer]:Cert & 2.403 & -1.103 & 5.910 & 0.179 &  & 4.522 \\
QCat[T.religion]:Cert & -0.851 & -2.137 & 0.434 & 0.194 &  & 1.268 \\
QCat[T.screenwriter]:Cert & -3.343 & -8.438 & 1.753 & 0.199 &  & -1.223 \\
QCat[T.sport]:Cert & -0.055 & -1.271 & 1.161 & 0.930 &  & 2.064 \\
\end{tabular}
\caption{Logistic regression results for model MT0-XXL.
\label{tbl:logistic_regression_mt0_xxl}}
\end{table*}

\begin{table*}[hbt]
\small
\centering
\begin{tabular}{l|rrrrlr}
 & coefficient ($\hat{\beta}$)& [0.025 & 0.975] & p-value & symbol & conditional coefficient \\
\hline
intercept & -1.678 & -1.900 & -1.457 & 0.000 & *** & -1.678 \\
QCat[T.capital] & -0.885 & -1.613 & -0.156 & 0.017 & * & -2.563 \\
QCat[T.capital of] & 0.317 & -0.248 & 0.883 & 0.271 &  & -1.361 \\
QCat[T.color] & -0.195 & -2.309 & 1.920 & 0.857 &  & -1.873 \\
QCat[T.composer] & -0.438 & -0.786 & -0.091 & 0.013 & * & -2.117 \\
QCat[T.country] & 1.979 & 1.595 & 2.362 & 0.000 & *** & 0.300 \\
QCat[T.director] & -2.233 & -2.708 & -1.759 & 0.000 & *** & -3.912 \\
QCat[T.father] & -0.714 & -1.178 & -0.250 & 0.003 & ** & -2.392 \\
QCat[T.genre] & -0.826 & -1.165 & -0.488 & 0.000 & *** & -2.505 \\
QCat[T.mother] & -2.124 & -3.200 & -1.048 & 0.000 & *** & -3.802 \\
QCat[T.occupation] & 0.548 & 0.187 & 0.908 & 0.003 & ** & -1.131 \\
QCat[T.place of birth] & 0.064 & -0.448 & 0.575 & 0.808 &  & -1.615 \\
QCat[T.producer] & -2.002 & -2.577 & -1.428 & 0.000 & *** & -3.681 \\
QCat[T.religion] & -0.464 & -1.266 & 0.338 & 0.257 &  & -2.142 \\
QCat[T.screenwriter] & -1.020 & -1.497 & -0.544 & 0.000 & *** & -2.699 \\
QCat[T.sport] & 1.397 & 0.988 & 1.806 & 0.000 & *** & -0.282 \\
\hline
log\_SPop & 1.987 & 1.685 & 2.290 & 0.000 & *** & 1.987 \\
QCat[T.capital]:log\_SPop & -1.230 & -1.595 & -0.866 & 0.000 & *** & 0.757 \\
QCat[T.capital of]:log\_SPop & -0.603 & -1.048 & -0.157 & 0.008 & ** & 1.385 \\
QCat[T.color]:log\_SPop & -0.434 & -1.828 & 0.960 & 0.542 &  & 1.553 \\
QCat[T.composer]:log\_SPop & -0.800 & -1.203 & -0.398 & 0.000 & *** & 1.187 \\
QCat[T.country]:log\_SPop & -1.915 & -2.269 & -1.561 & 0.000 & *** & 0.072 \\
QCat[T.director]:log\_SPop & 1.062 & 0.563 & 1.561 & 0.000 & *** & 3.049 \\
QCat[T.father]:log\_SPop & -0.850 & -1.232 & -0.467 & 0.000 & *** & 1.138 \\
QCat[T.genre]:log\_SPop & -1.110 & -1.457 & -0.763 & 0.000 & *** & 0.877 \\
QCat[T.mother]:log\_SPop & -0.620 & -1.238 & -0.001 & 0.050 & * & 1.368 \\
QCat[T.occupation]:log\_SPop & -1.300 & -1.679 & -0.921 & 0.000 & *** & 0.688 \\
QCat[T.place of birth]:log\_SPop & -1.591 & -2.008 & -1.173 & 0.000 & *** & 0.397 \\
QCat[T.producer]:log\_SPop & 0.156 & -0.326 & 0.637 & 0.527 &  & 2.143 \\
QCat[T.religion]:log\_SPop & -1.532 & -1.947 & -1.116 & 0.000 & *** & 0.455 \\
QCat[T.screenwriter]:log\_SPop & 0.465 & 0.017 & 0.913 & 0.042 & * & 2.453 \\
QCat[T.sport]:log\_SPop & -1.428 & -1.811 & -1.046 & 0.000 & *** & 0.559 \\
\hline
SCons & 0.266 & 0.021 & 0.511 & 0.034 & * & 0.266 \\
QCat[T.capital]:SCons & 0.379 & -0.163 & 0.920 & 0.171 &  & 0.645 \\
QCat[T.capital of]:SCons & 0.122 & -0.276 & 0.519 & 0.549 &  & 0.388 \\
QCat[T.color]:SCons & -0.365 & -1.347 & 0.616 & 0.466 &  & -0.099 \\
QCat[T.composer]:SCons & -0.035 & -0.396 & 0.327 & 0.851 &  & 0.231 \\
QCat[T.country]:SCons & -0.111 & -0.510 & 0.288 & 0.585 &  & 0.155 \\
QCat[T.director]:SCons & 0.464 & 0.083 & 0.845 & 0.017 & * & 0.730 \\
QCat[T.father]:SCons & 0.257 & -0.158 & 0.671 & 0.225 &  & 0.523 \\
QCat[T.genre]:SCons & 0.743 & 0.358 & 1.128 & 0.000 & *** & 1.009 \\
QCat[T.mother]:SCons & -0.158 & -0.980 & 0.665 & 0.707 &  & 0.108 \\
QCat[T.occupation]:SCons & -0.079 & -0.430 & 0.272 & 0.659 &  & 0.187 \\
QCat[T.place of birth]:SCons & -0.250 & -0.733 & 0.232 & 0.309 &  & 0.015 \\
QCat[T.producer]:SCons & 0.468 & 0.078 & 0.858 & 0.019 & * & 0.734 \\
QCat[T.religion]:SCons & -0.215 & -0.857 & 0.428 & 0.512 &  & 0.051 \\
QCat[T.screenwriter]:SCons & 0.845 & 0.447 & 1.243 & 0.000 & *** & 1.111 \\
QCat[T.sport]:SCons & 0.620 & 0.223 & 1.017 & 0.002 & ** & 0.886 \\
\hline
Cert & 1.326 & 1.030 & 1.622 & 0.000 & *** & 1.326 \\
QCat[T.capital]:Cert & 0.556 & -0.037 & 1.150 & 0.066 & . & 1.883 \\
QCat[T.capital of]:Cert & -1.078 & -1.607 & -0.549 & 0.000 & *** & 0.248 \\
QCat[T.color]:Cert & -0.114 & -1.520 & 1.293 & 0.874 &  & 1.213 \\
QCat[T.composer]:Cert & -0.283 & -0.704 & 0.138 & 0.187 &  & 1.043 \\
QCat[T.country]:Cert & -0.741 & -1.134 & -0.348 & 0.000 & *** & 0.585 \\
QCat[T.director]:Cert & -0.928 & -1.317 & -0.538 & 0.000 & *** & 0.398 \\
QCat[T.father]:Cert & -0.382 & -0.886 & 0.122 & 0.137 &  & 0.944 \\
QCat[T.genre]:Cert & -1.030 & -1.439 & -0.620 & 0.000 & *** & 0.296 \\
QCat[T.mother]:Cert & 0.392 & -0.587 & 1.370 & 0.433 &  & 1.718 \\
QCat[T.occupation]:Cert & -1.280 & -1.698 & -0.863 & 0.000 & *** & 0.046 \\
QCat[T.place of birth]:Cert & -0.869 & -1.320 & -0.417 & 0.000 & *** & 0.457 \\
QCat[T.producer]:Cert & -1.050 & -1.478 & -0.623 & 0.000 & *** & 0.276 \\
QCat[T.religion]:Cert & -1.656 & -2.350 & -0.963 & 0.000 & *** & -0.330 \\
QCat[T.screenwriter]:Cert & -0.695 & -1.096 & -0.295 & 0.001 & *** & 0.631 \\
QCat[T.sport]:Cert & -1.356 & -1.781 & -0.932 & 0.000 & *** & -0.030 \\
\end{tabular}
\caption{Logistic regression results for model MPT-Instruct2.
\label{tbl:logistic_regression_mpt_instruct2}}
\end{table*}

\begin{table*}[hbt]
\small
\centering
\begin{tabular}{l|rrrrlr}
 & coefficient ($\hat{\beta}$)& [0.025 & 0.975] & p-value & symbol & conditional coefficient \\
\hline
intercept & -2.505 & -2.767 & -2.243 & 0.000 & *** & -2.505 \\
QCat[T.capital] & 3.063 & 2.459 & 3.668 & 0.000 & *** & 0.559 \\
QCat[T.capital of] & 0.570 & -0.202 & 1.341 & 0.148 &  & -1.935 \\
QCat[T.color] & 2.552 & 0.791 & 4.314 & 0.005 & ** & 0.048 \\
QCat[T.composer] & -0.017 & -0.412 & 0.378 & 0.932 &  & -2.522 \\
QCat[T.country] & 3.126 & 2.754 & 3.499 & 0.000 & *** & 0.622 \\
QCat[T.director] & -2.483 & -3.201 & -1.765 & 0.000 & *** & -4.988 \\
QCat[T.father] & 0.574 & 0.118 & 1.031 & 0.014 & * & -1.930 \\
QCat[T.genre] & 0.262 & -0.079 & 0.603 & 0.132 &  & -2.243 \\
QCat[T.mother] & -0.873 & -2.543 & 0.797 & 0.305 &  & -3.378 \\
QCat[T.occupation] & -0.049 & -0.586 & 0.488 & 0.858 &  & -2.554 \\
QCat[T.place of birth] & 0.666 & 0.237 & 1.094 & 0.002 & ** & -1.839 \\
QCat[T.producer] & -2.042 & -2.708 & -1.375 & 0.000 & *** & -4.546 \\
QCat[T.religion] & 1.184 & 0.459 & 1.910 & 0.001 & ** & -1.320 \\
QCat[T.screenwriter] & -1.264 & -1.793 & -0.734 & 0.000 & *** & -3.769 \\
QCat[T.sport] & 2.843 & 2.421 & 3.265 & 0.000 & *** & 0.338 \\
\hline
log\_SPop & 2.072 & 1.763 & 2.382 & 0.000 & *** & 2.072 \\
QCat[T.capital]:log\_SPop & -1.307 & -1.670 & -0.944 & 0.000 & *** & 0.765 \\
QCat[T.capital of]:log\_SPop & -0.834 & -1.253 & -0.416 & 0.000 & *** & 1.238 \\
QCat[T.color]:log\_SPop & -1.687 & -2.776 & -0.598 & 0.002 & ** & 0.385 \\
QCat[T.composer]:log\_SPop & -0.989 & -1.408 & -0.571 & 0.000 & *** & 1.083 \\
QCat[T.country]:log\_SPop & -1.937 & -2.292 & -1.582 & 0.000 & *** & 0.135 \\
QCat[T.director]:log\_SPop & 0.783 & 0.189 & 1.378 & 0.010 & ** & 2.855 \\
QCat[T.father]:log\_SPop & -1.344 & -1.716 & -0.972 & 0.000 & *** & 0.728 \\
QCat[T.genre]:log\_SPop & -1.243 & -1.599 & -0.887 & 0.000 & *** & 0.829 \\
QCat[T.mother]:log\_SPop & -0.854 & -1.532 & -0.177 & 0.013 & * & 1.218 \\
QCat[T.occupation]:log\_SPop & -1.465 & -1.889 & -1.040 & 0.000 & *** & 0.608 \\
QCat[T.place of birth]:log\_SPop & -1.836 & -2.278 & -1.394 & 0.000 & *** & 0.237 \\
QCat[T.producer]:log\_SPop & 0.068 & -0.453 & 0.589 & 0.798 &  & 2.140 \\
QCat[T.religion]:log\_SPop & -1.956 & -2.347 & -1.566 & 0.000 & *** & 0.116 \\
QCat[T.screenwriter]:log\_SPop & 0.162 & -0.296 & 0.621 & 0.488 &  & 2.234 \\
QCat[T.sport]:log\_SPop & -1.960 & -2.335 & -1.585 & 0.000 & *** & 0.112 \\
\hline
SCons & 0.172 & -0.048 & 0.392 & 0.125 &  & 0.172 \\
QCat[T.capital]:SCons & 0.670 & 0.270 & 1.071 & 0.001 & ** & 0.842 \\
QCat[T.capital of]:SCons & 0.040 & -0.325 & 0.405 & 0.828 &  & 0.212 \\
QCat[T.color]:SCons & 0.698 & -0.470 & 1.867 & 0.241 &  & 0.870 \\
QCat[T.composer]:SCons & 0.173 & -0.181 & 0.527 & 0.339 &  & 0.345 \\
QCat[T.country]:SCons & -0.404 & -0.694 & -0.113 & 0.006 & ** & -0.232 \\
QCat[T.director]:SCons & 0.318 & -0.111 & 0.747 & 0.146 &  & 0.490 \\
QCat[T.father]:SCons & 0.173 & -0.141 & 0.488 & 0.280 &  & 0.345 \\
QCat[T.genre]:SCons & 0.153 & -0.171 & 0.478 & 0.354 &  & 0.325 \\
QCat[T.mother]:SCons & 0.463 & -0.254 & 1.180 & 0.206 &  & 0.635 \\
QCat[T.occupation]:SCons & 0.025 & -0.386 & 0.436 & 0.906 &  & 0.197 \\
QCat[T.place of birth]:SCons & -0.182 & -0.530 & 0.166 & 0.304 &  & -0.010 \\
QCat[T.producer]:SCons & -0.341 & -0.754 & 0.071 & 0.105 &  & -0.169 \\
QCat[T.religion]:SCons & -0.208 & -0.594 & 0.178 & 0.291 &  & -0.036 \\
QCat[T.screenwriter]:SCons & 0.498 & 0.130 & 0.865 & 0.008 & ** & 0.670 \\
QCat[T.sport]:SCons & -0.292 & -0.593 & 0.009 & 0.057 & . & -0.120 \\
\hline
Cert & 0.520 & 0.258 & 0.782 & 0.000 & *** & 0.520 \\
QCat[T.capital]:Cert & 0.738 & 0.280 & 1.195 & 0.002 & ** & 1.258 \\
QCat[T.capital of]:Cert & -0.094 & -0.518 & 0.329 & 0.662 &  & 0.426 \\
QCat[T.color]:Cert & -0.518 & -1.714 & 0.678 & 0.396 &  & 0.002 \\
QCat[T.composer]:Cert & 0.207 & -0.206 & 0.620 & 0.325 &  & 0.727 \\
QCat[T.country]:Cert & -0.077 & -0.398 & 0.244 & 0.638 &  & 0.443 \\
QCat[T.director]:Cert & 0.378 & -0.120 & 0.876 & 0.137 &  & 0.898 \\
QCat[T.father]:Cert & -0.415 & -0.791 & -0.040 & 0.030 & * & 0.105 \\
QCat[T.genre]:Cert & -0.243 & -0.590 & 0.103 & 0.169 &  & 0.277 \\
QCat[T.mother]:Cert & -0.203 & -1.521 & 1.115 & 0.763 &  & 0.317 \\
QCat[T.occupation]:Cert & -0.284 & -0.813 & 0.246 & 0.294 &  & 0.237 \\
QCat[T.place of birth]:Cert & -0.005 & -0.425 & 0.416 & 0.983 &  & 0.515 \\
QCat[T.producer]:Cert & 0.378 & -0.102 & 0.858 & 0.123 &  & 0.898 \\
QCat[T.religion]:Cert & -0.710 & -1.187 & -0.232 & 0.004 & ** & -0.190 \\
QCat[T.screenwriter]:Cert & 0.486 & 0.034 & 0.938 & 0.035 & * & 1.006 \\
QCat[T.sport]:Cert & -0.266 & -0.610 & 0.077 & 0.129 &  & 0.254 \\
\end{tabular}
\caption{Logistic regression results for model GPT-NeoX-20B.
\label{tbl:logistic_regression_gpt_neox}}
\end{table*}

\end{document}